\documentclass[sigconf,nonacm]{acmart}

\author{Minhee Park}
\email{minheepark@hanyang.ac.kr}
\affiliation{%
  \institution{Hanyang University}
  \city{Seoul}
  \country{Republic of Korea}
}

\author{Seongyeon Son}
\email{annssy@hanyang.ac.kr}
\affiliation{%
  \institution{Hanyang University}
  \city{Seoul}
  \country{Republic of Korea}
}

\author{Yonghyun Lee}
\email{yhlee@kmu.ac.kr}
\affiliation{%
  \institution{Keimyung University}
  \city{Daegu}
  \country{Republic of Korea}
}

\author{Eunchan Kim}
\email{eckim@hanyang.ac.kr}
\affiliation{%
  \institution{Hanyang University}
  \city{Seoul}
  \country{Republic of Korea}
}

\AtBeginDocument{%
  }

\setcopyright{none}
\renewcommand\footnotetextcopyrightpermission[1]{}

\usepackage{algorithm}
\usepackage{algorithmic}
\usepackage{graphicx}
\usepackage{subcaption}
\usepackage{placeins}
\usepackage{array}

\hypersetup{
    pdftitle={SCOPE-FE: Structured Control of Operator and Pairwise Exploration for Feature Engineering via Quality-Aware Candidate-Space Reduction},
    pdfauthor={Minhee Park, Seongyeon Son, Yonghyun Lee, and Eunchan Kim},
    pdfkeywords={Automated Feature Engineering, Feature Generation, Search-Space Control, Candidate-Space Reduction, Tabular Learning}
}

\begin{document}

\title[SCOPE-FE: Operator and Pairwise Search-Space Control]{SCOPE-FE: Structured Control of Operator and Pairwise Exploration for Feature Engineering via Quality-Aware Candidate-Space Reduction}


\renewcommand{\shortauthors}{Park et al.}

\begin{abstract}
    Automatic feature engineering can improve predictive performance on tabular data by generating diverse feature transformations. However, the candidate space induced by combinations of input features and operators grows rapidly with dimensionality, resulting in substantial computational cost. We propose SCOPE-FE, a framework that controls the search space before candidate generation. SCOPE-FE combines FeatureClustering, a structural pair gate based on mixed-type feature association, with OperatorProbing, a dataset-specific utility control over operators. Unlike conventional expand-and-reduce approaches that generate a large candidate set and prune it afterward, SCOPE-FE focuses computation on a smaller, data-dependent candidate pool. Across ten OpenFE benchmark datasets, SCOPE-FE achieves a median candidate-space reduction of 82.9\% and lowers component-summed feature-engineering time---including separately measured FeatureClustering overhead---on all ten datasets, yielding a geometric-mean speedup of 2.66$\times$ and a maximum speedup of 5.48$\times$. Despite this reduction, SCOPE-FE is within the stated practical-equivalence margin of OpenFE on 8 of 10 datasets. An exhaustive candidate audit shows enrichment above uniform-random expectation on 8 of 10 datasets, with a median enrichment of 1.35$\times$. Against Random-Pair, SCOPE-FE has higher enrichment on 6 of 10 datasets, with 5 of 10 significant; against Random-Operator and Random-Joint, it has higher enrichment on 8 of 10 datasets, with 8 of 10 significant for each. These results demonstrate that pre-generation search-space control can substantially reduce feature-engineering time while retaining a utility-enriched candidate pool under the OpenFE-compatible evaluation protocol.
\end{abstract}

\ccsdesc[500]{Computing methodologies~Feature selection}
\ccsdesc[300]{Information systems~Data mining}
\ccsdesc[100]{Computing methodologies~Supervised learning}

\keywords{Automated Feature Engineering, Feature Generation, Search-Space Control, Candidate-Space Reduction, Tabular Learning}


\maketitle

\section{Introduction}
\label{sec:introduction}

Feature engineering remains central to machine learning on tabular data because transformations of the original variables can expose predictive structure that is not directly represented by the raw columns \cite{ref1,ref5,ref7}. Recent benchmarks further show that predictive performance on tabular data varies materially across model families and their inductive biases \cite{gorishniy2021tabular,grinsztajn2022trees,shwartz2022tabular}. Automatic feature engineering (AutoFE) seeks to reduce the manual effort required for this process by systematically constructing candidate features from predefined transformation operators. Many AutoFE systems use generation-and-selection or expand-and-reduce pipelines \cite{ref1,ref5,ref7,ref12,ref14}. OpenFE \cite{ref7}, for example, generates unary and two-input transformations and progressively evaluates them using FeatureBoost, successive halving, and feature attribution.

Although effective, this paradigm can incur substantial computational cost because two-input transformations are applied across feature pairs, causing the candidate space to grow quadratically with input dimensionality \cite{ref7,ref14}. High-dimensional datasets may therefore require generating and evaluating many candidates even though only a small subset is ultimately selected.

Prior work reduces this burden through pruning, ranking, and iterative selection \cite{ref1,ref7,ref11}; beam search \cite{ref17}; reinforcement learning \cite{ref18,ref19}; neural feature search \cite{ref20}; and differentiable optimization \cite{ref10}. Related work on adaptive resource allocation includes successive halving, Hyperband, and BOHB \cite{jamieson2016successive,li2018hyperband,falkner2018bohb}. However, many of these approaches guide exploration or allocate evaluation after a broad search space has been specified. Related work has also coupled columns with transformation proposals through foundation models \cite{ref31}, or learned feature-interaction structure within specialized predictive or search models \cite{liu2020autofis,xie2021fives}. This motivates a complementary question: can the feature-pair and operator spaces be explicitly factorized and controlled \emph{before} full candidate generation while retaining an otherwise unchanged downstream selector?

We propose \textbf{SCOPE-FE}, a framework that controls the search space before candidate generation. \textbf{FeatureClustering} acts as a structural pair gate: it constructs a mixed-type feature-association matrix, applies deterministic agglomerative clustering, and restricts two-input transformations to intra-cluster pairs. \textbf{OperatorProbing} provides dataset-specific utility control through lightweight evaluations on subsampled development data under the OpenFE-compatible cross-fitted candidate-selection protocol, retaining only the highest-scoring operators for full candidate generation. Together, the two components provide joint quality-aware candidate-space control before the standard OpenFE selection pipeline is applied. SCOPE-FE thus makes the pair and operator scopes explicit while leaving downstream candidate scoring and feature selection unchanged. We refer to this process as quality-aware candidate-space reduction because pair admission is guided by mixed-type feature association and operator admission is guided by dataset-specific utility, rather than by uniform or arbitrary removal.

We evaluate SCOPE-FE on ten datasets from the OpenFE benchmark suite, covering regression, binary classification, and multiclass classification with widely varying sample sizes and dimensionalities. SCOPE-FE achieves a median candidate-space reduction of 82.9\%, decreases component-summed RUN on all ten datasets, and achieves a geometric-mean speedup of 2.66$\times$ with a maximum speedup of 5.48$\times$. Under the defined practical-equivalence margin, it remains comparable to full OpenFE on 8 of 10 datasets and incurs losses on California Housing and Telecom. An exhaustive candidate audit shows enrichment above uniform-random expectation on 8 of 10 datasets, with a median enrichment of 1.35$\times$; these results remain significant after Benjamini--Hochberg correction on the same 8 datasets. Against Random-Pair, SCOPE-FE has higher enrichment on 6 of 10 datasets, with 5 of 10 significant; against Random-Operator and Random-Joint, it has higher enrichment on 8 of 10 datasets, with 8 of 10 significant for each.

SCOPE-FE contributes a factorized formulation of pre-generation scope control, in which feature-pair admission and operator admission constitute independently controllable axes. A structural pair gate and a dataset-specific utility gate are jointly applied before an otherwise unchanged OpenFE evaluation pipeline, enabling the effects of pair control, operator control, and their combination to be isolated directly.

Our main contributions are as follows:
\begin{itemize}
    \item We formulate AutoFE as controlling the search space before generation, thereby reducing candidate generation and evaluation costs before full expansion.

    \item We introduce SCOPE-FE, a factorized pair--operator control framework that combines deterministic mixed-type FeatureClustering as a structural pair gate with lightweight OperatorProbing as dataset-specific utility control.

    \item We show that the retained candidate pool is utility-enriched on most evaluated datasets and that this enrichment is not explained by candidate count alone, using exhaustive candidate-universe auditing, budget-matched randomization controls, and component ablations.
\end{itemize}

\section{Related Work}
\label{sec:related_work}

\subsection{Automated Feature Construction and Search Control}

Automated feature construction spans compositional synthesis, supervised search, policy learning, and context-aware generation. Deep Feature Synthesis composes transformation primitives over relational data \cite{kanter2015deep}, while ExploreKit, AutoLearn, and Cognito construct and evaluate transformations for supervised tasks \cite{ref1,ref5,ref12}. AutoFeat uses a multi-step generation-and-selection process to construct nonlinear features while retaining a compact selected subset \cite{ref11}. Learning Feature Engineering and recent AutoFE--AutoML work learn from prior datasets to recommend transformations or operators without explicitly expanding the full candidate space \cite{ref6,ref4}. External and semantic knowledge have also been incorporated through DBpedia-based generation \cite{ref2} and large-language-model context in CAAFE \cite{hollmann2023caafe}.

Representative explicit construction systems use different mechanisms to limit exploration. FCTree applies divide-and-conquer local feature construction and evaluation in error-prone example subspaces \cite{ref15}. SAFE mines promising feature combinations from boosting-tree paths before applying predefined operators \cite{ref16}; AutoCross searches high-order categorical crosses with beam search \cite{ref17}; and FETCH learns a transferable feature-engineering policy \cite{ref3}. OpenFE combines a general operator library with FeatureBoost-based evaluation and coarse-to-fine pruning \cite{ref7}. More recent pair-oriented methods narrow construction before broad enumeration: OpenFE++ uses local feature interactions for tabular data and representative lagged periods for time-series data to construct a reduced candidate pool \cite{ref14}, whereas IIFE ranks synergistic feature pairs through interaction information and uses them to guide feature construction \cite{overman2024iife}.

The combinatorial cost of feature construction has been addressed through beam search \cite{ref17}, reinforcement learning \cite{ref18,ref19}, neural feature search or generation \cite{ref20,ref21}, and differentiable optimization \cite{ref10}. Related resource-allocation methods progressively eliminate weak configurations or vary evaluation budgets across alternatives \cite{jamieson2016successive,li2018hyperband,falkner2018bohb}. These methods reduce optimization or evaluation expenditure over admitted alternatives. SCOPE-FE instead changes the explicit pair--operator scope from which the full OpenFE candidate pool is materialized, making pre-generation control complementary to progressive evaluation.

\subsection{Feature Interactions and Structured Pair Control}

Feature interactions have been modeled through factorization, explicit deep crosses, attention, and interaction-structure search \cite{rendle2010fm,wang2017dcn,song2019autoint,liu2020autofis,xie2021fives}. These approaches primarily optimize interactions within specialized predictive or search architectures rather than controlling an OpenFE-style explicit transformation pool.

Pair control also draws on feature selection and grouping. Classical work formalizes relevance, redundancy, and mutual-information criteria \cite{guyon2003feature,yu2004relevance,peng2005mrmr}, while feature and hierarchical clustering organize related variables \cite{dhillon2003featureclustering,murtagh2012hierarchical}. Dependency-aware and pairwise selection prioritize features or feature pairs \cite{ref22,ref23}; general clustering provides the algorithmic background \cite{ref24,ref25}; graph-based feature grouping explicitly organizes related variables \cite{ref26}; and interaction analysis characterizes nonadditive effects \cite{ref27}. However, these selection, clustering, and interaction-analysis methods do not define a joint pair--operator gate for explicit AutoFE.

Several AutoFE systems already couple feature and operator decisions to varying degrees. SAFE identifies feature combinations before operator application \cite{ref16}; CN2-MCI and IIFE select feature pairs prior to construction \cite{ref32,overman2024iife}; OpenFE++ restricts interactions used for candidate generation \cite{ref14}; and SMARTFEAT uses a foundation-model-guided selector to propose operators together with relevant columns \cite{ref31}. Relative to these methods, SCOPE-FE explicitly factorizes pre-generation scope control into a deterministic structural pair gate and lightweight dataset-specific utility control, applies both before an otherwise unchanged OpenFE scoring and selection pipeline, and evaluates the retained pool through an exhaustive candidate audit.

\begin{figure*}[!t]
    \centering
    \includegraphics[width=\textwidth]{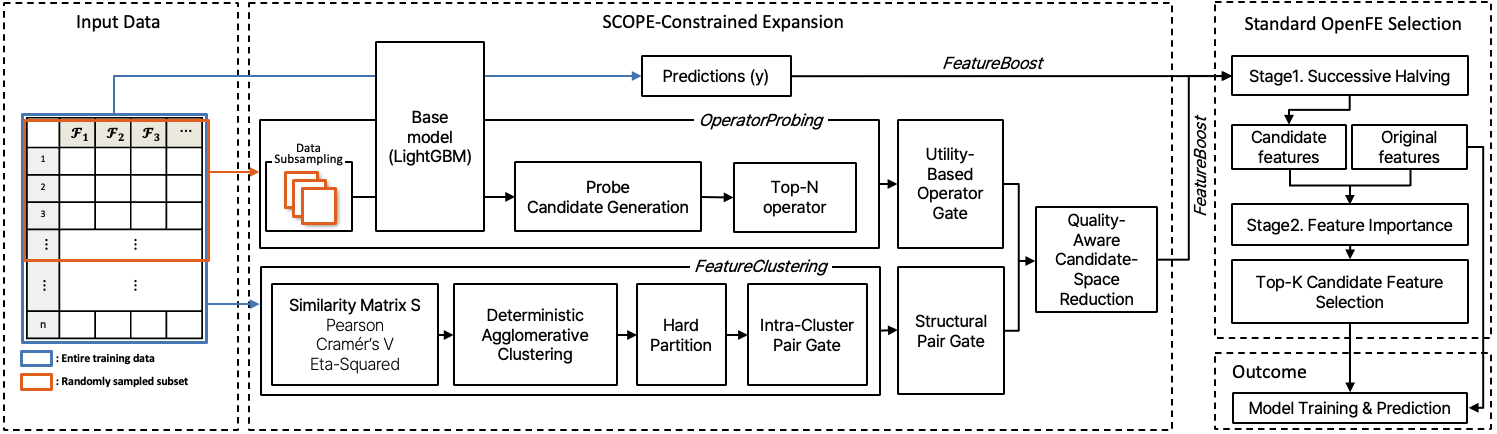}
    \caption{Overview of SCOPE-FE. FeatureClustering provides a structural pair gate by constructing a deterministic hard partition from mixed-type feature association and restricting two-input transformations to intra-cluster feature pairs. OperatorProbing provides dataset-specific utility control through lightweight subsampled evaluations. Their combination provides joint quality-aware candidate-space control, after which the reduced pool is processed using the standard OpenFE selection pipeline.}
    \Description{A full-width pipeline diagram showing the FeatureClustering and OperatorProbing branches of SCOPE-FE, followed by constrained candidate generation, successive halving, feature attribution, and final feature selection.}
    \label{fig:scope_overview}
\end{figure*}

\section{Problem Formulation}
\label{sec:problem_formulation}

Given a supervised learning dataset $D=\{(\mathbf{x}_i,y_i)\mid i=1,\ldots,n\}$, partitioned into a training set $D_{\mathrm{tr}}$ and an internal candidate-selection set $D_{\mathrm{sel}}$ (the original OpenFE validation split), let $T=\{t_1,\ldots,t_d\}$ denote the set of original features and $O=\{o_1,\ldots,o_p\}$ the set of transformation operators. Together, $D_{\mathrm{tr}}$ and $D_{\mathrm{sel}}$ form the development data used by the OpenFE-compatible cross-fitted candidate-selection protocol. We define the full candidate universe as
\begin{equation}
    \mathcal{U}=\mathcal{A}(T,O),
    \label{eq:full_candidate_universe}
\end{equation}
where $\mathcal{A}(T,O)$ contains all valid features constructible by applying the operators in $O$ to the features in $T$. Equation~\eqref{eq:full_candidate_universe} serves as the reference universe for the reduced pools defined below.

Let $L$ denote the downstream learning algorithm, and let $k_{\mathrm{sel}}$ be the maximum number of generated features retained for final model training. We define $E$ as an evaluation utility oriented such that larger values are better, using the negative of error metrics such as RMSE when necessary. For a candidate pool $\mathcal{B}\subseteq\mathcal{U}$, we define its ideal achievable downstream performance as
\begin{equation}
\begin{aligned}
    P(\mathcal{B})
    =
    \max_{\substack{
        T_{\mathrm{new}}\subseteq\mathcal{B}\\
        |T_{\mathrm{new}}|\leq k_{\mathrm{sel}}
    }}
    E\Big(
        L(D_{\mathrm{tr}},T\cup T_{\mathrm{new}}),\,
        D_{\mathrm{sel}}
    \Big).
\end{aligned}
\label{eq:pool_performance}
\end{equation}
Here, $L(D_{\mathrm{tr}},T\cup T_{\mathrm{new}})$ denotes a model trained using the augmented feature set, and $E(\cdot,D_{\mathrm{sel}})$ denotes the internal selection utility used to compare candidate feature sets. This quantity is a development-stage selection objective, not a final model-evaluation estimate. Expand-and-reduce AutoFE methods such as OpenFE seek to approximate Eq.~\eqref{eq:pool_performance} over a broadly defined candidate universe, corresponding to $\mathcal{B}=\mathcal{U}$.

OpenFE estimates the incremental utility of an individual candidate $\phi\in\mathcal{U}$ using FeatureBoost. Given the initial internal-selection loss $l_{\mathrm{init}}$ and the best internal-selection loss $l_{\mathrm{best}}$ obtained when evaluating $\phi$, its individual gain is defined as
\begin{equation}
    \Delta(\phi)
    =
    l_{\mathrm{init}}-l_{\mathrm{best}}.
    \label{eq:featureboost}
\end{equation}
A positive value of $\Delta(\phi)$ indicates that the candidate improves the internal selection objective under the FeatureBoost evaluation. This candidate-level utility is used for progressive evaluation and selection within OpenFE.

The difficulty is that the full candidate universe grows rapidly with the number of input features. Let $O_1$ and $O_2$ denote the sets of one-input and two-input operators, respectively. Ignoring type constraints and lower-order terms, the candidate-space size grows as
\begin{equation}
    |\mathcal{U}|
    =
    \mathcal{O}\bigl(
        |O_1|d+|O_2|d^2
    \bigr)
    =
    \mathcal{O}(pd^2).
    \label{eq:searchspace}
\end{equation}
As shown in Eq.~\eqref{eq:searchspace}, the quadratic term is induced by two-input transformations applied over feature pairs. Moreover, each generated candidate requires candidate-level FeatureBoost evaluation. Consequently, the computational cost depends not only on the features ultimately selected but also on all candidates generated, evaluated, and subsequently discarded.

We therefore formulate efficient AutoFE as a \emph{pre-generation search-space control} problem. A scope-control function $G$ constructs a reduced candidate pool
\begin{equation}
    A
    =
    G(D,T,O)
    \subseteq
    \mathcal{U},
    \label{eq:reduced_candidate_pool}
\end{equation}
before the full candidate-evaluation procedure is executed. The reduced-pool relation in Eq.~\eqref{eq:reduced_candidate_pool} makes candidate generation itself part of the controlled scope. Ideally, $A$ should substantially reduce computational cost while preserving the downstream utility available in the full universe. This trade-off can be expressed through the following idealized constrained objective:
\begin{equation}
\begin{aligned}
    \min_G \quad&
    \mathrm{Cost}(A)\\
    \text{subject to}\quad&
    P(A)
    \geq
    P(\mathcal{U})-\epsilon,
\end{aligned}
\label{eq:scope_control_objective}
\end{equation}
where $\mathrm{Cost}(A)$ includes the costs of constructing, generating, and evaluating the retained candidate pool, and $\epsilon\geq0$ denotes an acceptable downstream-performance difference.

Directly solving Eq.~\eqref{eq:scope_control_objective} would require knowledge of the utility available throughout the full candidate universe and would therefore defeat the purpose of search-space control. SCOPE-FE instead constructs $A$ using two lightweight, dataset-dependent controls: an operator subset $O^*\subseteq O$ selected by OperatorProbing and a hard cluster-assignment function $C_{\mathrm{hard}}$ constructed by FeatureClustering. The resulting candidate pool is
\begin{equation}
    A
    =
    \mathcal{A}(T,O^*,C_{\mathrm{hard}})
    \subseteq
    \mathcal{A}(T,O).
    \label{eq:scope_candidate_space}
\end{equation}
The specific intra-cluster pairing rule induced by $C_{\mathrm{hard}}$ is described in Section~\ref{sec:scope_fe}. The standard OpenFE selection procedure is then applied only to $A$.

A small candidate pool is not necessarily a useful one. Search-space control should therefore be evaluated not only by its reduction ratio and final downstream performance but also by the utility composition of the retained pool. In particular, we examine whether $A$ contains a higher concentration of high-individual-gain candidates than an equally sized subset drawn uniformly from $\mathcal{U}$. We use coverage and enrichment as post-hoc evaluation metrics for this purpose, as formalized in Section~\ref{sec:candidate_quality_metrics}.

\section{SCOPE-FE}
\label{sec:scope_fe}

SCOPE-FE instantiates the scope-control function $G$ introduced in Section~\ref{sec:problem_formulation} through two complementary, dataset-dependent mechanisms. FeatureClustering provides a structural pair gate by constructing a deterministic hard partition of the original features from mixed-type feature association and allowing two-input transformations only between features assigned to the same cluster. OperatorProbing provides dataset-specific utility control through lightweight subsampled evaluations and retains only a compact subset of operators. Their combination provides joint quality-aware candidate-space control. The resulting reduced candidate pool $A=\mathcal{A}(T,O^*,C_{\mathrm{hard}})$ is constructed before full candidate evaluation, after which the standard OpenFE successive-halving and feature-attribution procedures are applied without modification.

\subsection{Overall Architecture}
\label{subsec:overall_architecture}
The SCOPE-FE pipeline consists of four stages. 
Figure~\ref{fig:scope_overview} illustrates the complete pipeline, and Algorithm~\ref{alg:scope_openfe} summarizes the corresponding procedure. First, FeatureClustering implements the structural pair gate by constructing a mixed-type feature-association matrix and partitioning the original features using deterministic agglomerative clustering. Second, OperatorProbing implements dataset-specific utility control by evaluating a small number of sampled candidates for each operator and selecting the highest-scoring operators. Third, constrained candidate generation applies the selected operators while restricting two-input transformations to intra-cluster feature pairs. Finally, the resulting candidate pool is processed by the standard OpenFE selection pipeline.

\begin{algorithm}[t]
    \caption{Space-Controlled OpenFE}
    \label{alg:scope_openfe}
    \begin{algorithmic}
        \REQUIRE $D=(D_{\mathrm{tr}},D_{\mathrm{sel}})$: training and internal candidate-selection data, $T$: feature set, $O$: operator set, $\tau$: target cluster size, $k_{\mathrm{sel}}$: maximum number of selected generated features
        \ENSURE $T \cup T_{\mathrm{new}}$: augmented feature set

        \STATE \textbf{Stage A: Feature Clustering}
        \STATE $C_{\mathrm{hard}} \leftarrow \text{FeatureClustering}(D_{\mathrm{tr}},T,\tau)$

        \STATE \textbf{Stage B: Operator Probing}
        \STATE $O^* \leftarrow \text{OperatorProbing}(D_{\mathrm{tr}},D_{\mathrm{sel}},T,O)$

        \STATE \textbf{Stage C: Constrained Candidate Generation}
        \STATE $A \leftarrow \text{GenerateCandidates}(T,O^*,C_{\mathrm{hard}})$

        \STATE \textbf{Stage D: Standard OpenFE Selection}
        \STATE $\hat{y} \leftarrow \text{BaselinePrediction}(D,T)$
        \STATE $A_{\mathrm{selected}} \leftarrow \text{SuccessiveHalving}(A,D,\hat{y})$
        \STATE $A_{\mathrm{final}} \leftarrow \text{FeatureAttribution}(A_{\mathrm{selected}},D,\hat{y})$
        \STATE $T_{\mathrm{new}} \leftarrow \text{TopK}(A_{\mathrm{final}},k_{\mathrm{sel}})$
        \STATE \textbf{return} $T \cup T_{\mathrm{new}}$
    \end{algorithmic}
\end{algorithm}

\subsection{Pair-Space Scope Control}
\label{subsec:pair_space}

OpenFE constructs two-input candidates by applying transformation operators to compatible feature pairs. Because the number of possible pairs grows quadratically with the number of original features, this expansion dominates the candidate space for sufficiently high-dimensional datasets. FeatureClustering acts as a structural pair gate by grouping statistically associated features and retaining only intra-cluster pairs. The clustering procedure is deterministic and assigns each feature to exactly one cluster.

\subsubsection{Mixed-Type Feature Association}
\label{subsubsec:similarity}

Using only the training partition $D_{\mathrm{tr}}$, we construct a symmetric similarity matrix $S\in[0,1]^{d\times d}$ for the $d$ original features, where $S_{ij}$ measures the statistical association between features $t_i$ and $t_j$. Because tabular datasets may contain both numerical and categorical features, the association measure depends on the feature-type pair:
\begin{equation}
S_{ij}=
\begin{cases}
\left|\rho(t_i,t_j)\right|, & t_i,t_j \text{ are numerical},\\
V_{\mathrm{corr}}(t_i,t_j), & t_i,t_j \text{ are categorical},\\
\eta^2(t_{\mathrm{cat}},t_{\mathrm{num}}), & \text{otherwise},
\end{cases}
\label{eq:mixed_similarity}
\end{equation}
where $\rho$ denotes Pearson correlation, $V_{\mathrm{corr}}$ denotes bias-corrected Cramér's V \cite{bergsma2013cramerv}, and $\eta^2$ denotes eta-squared. We set $S_{ii}=1$ and symmetrize the off-diagonal entries. Formal definitions of the three association statistics are provided in Appendix~\ref{app:similarity_measures}.

\subsubsection{Deterministic Agglomerative FeatureClustering}
\label{subsubsec:hard_clustering}

We convert the similarity matrix into a precomputed dissimilarity matrix
\begin{equation}
\widetilde{D}_{ij}=1-S_{ij}.
\label{eq:hard_distance}
\end{equation}
Let $\tau$ denote the target cluster size. The number of clusters is set to
\begin{equation}
K=\max\left(2,\left\lceil\frac{d}{\tau}\right\rceil\right).
\label{eq:num_clusters}
\end{equation}
The lower bound of two ensures a nontrivial partition and prevents the method from degenerating into the unconstrained single-cluster case when $d \leq \tau$. It does not guarantee that every cluster contains an admissible intra-cluster pair.
Agglomerative clustering is then applied directly to $\widetilde{D}$ to construct the hard cluster-assignment function and corresponding disjoint partition in Eq.~\eqref{eq:hard_partition}:
\begin{equation}
\begin{aligned}
C_{\mathrm{hard}} &: T\rightarrow\{1,\ldots,K\},\\
C_k &=
\{t_i\in T:C_{\mathrm{hard}}(t_i)=k\},\\
\bigcup_{k=1}^{K}C_k &=T,\qquad
C_k\cap C_{\ell}=\emptyset\quad(k\neq\ell).
\end{aligned}
\label{eq:hard_partition}
\end{equation}
FeatureClustering permits a two-input transformation only when both arguments belong to the same cluster:
\begin{equation}
I_{\mathrm{hard}}(t_i,t_j)=\mathbf{1}\!\left[C_{\mathrm{hard}}(t_i)=C_{\mathrm{hard}}(t_j)\right].
\label{eq:intra_indicator}
\end{equation}
Thus, the indicator in Eq.~\eqref{eq:intra_indicator} excludes cross-cluster pairs before candidate construction. Unary transformations are unaffected by the pair constraint, while two-input transformations additionally follow OpenFE's original type compatibility and argument ordering rules. Algorithm~\ref{alg:feature_clustering} summarizes the procedure.

\begin{algorithm}[t]
    \caption{Feature Clustering (Hard)}
    \label{alg:feature_clustering}
    \begin{algorithmic}
        \REQUIRE $D_{\mathrm{tr}}$: training data, $T=\{t_1,\ldots,t_d\}$: original features, $\tau$: target cluster size
        \ENSURE $C_{\mathrm{hard}}:T\rightarrow\{1,\ldots,K\}$: hard cluster assignment

        \STATE $K \leftarrow \max\left(2,\left\lceil |T|/\tau \right\rceil\right)$
        \STATE $S \leftarrow I_d$

        \STATE \textbf{for each pair} $(t_i,t_j)$ with $i<j$:
        \STATE \hspace*{1em} \textbf{if both numerical:} $S_{ij}\leftarrow|\rho(t_i,t_j)|$
        \STATE \hspace*{1em} \textbf{if both categorical:} $S_{ij}\leftarrow V_{\mathrm{corr}}(t_i,t_j)$
        \STATE \hspace*{1em} \textbf{if mixed:} $S_{ij}\leftarrow\eta^2(t_{\mathrm{cat}},t_{\mathrm{num}})$
        \STATE \hspace*{1em} $S_{ji}\leftarrow S_{ij}$

        \STATE $\widetilde{D} \leftarrow \mathbf{1}\mathbf{1}^{\top}-S$
        \STATE $C_{\mathrm{hard}} \leftarrow \text{AgglomerativeClustering}(\widetilde{D},K,\text{linkage}=\text{average})$
        \STATE \textbf{return} $C_{\mathrm{hard}}$
    \end{algorithmic}
\end{algorithm}

\subsection{Operator-Space Scope Control}
\label{subsec:operator_space}

OpenFE employs a fixed set of transformation operators (arithmetic, trigonometric, and aggregation). However, not all operators are informative for every dataset. For a dataset with $d$ features and $p$ operators, the number of binary candidates scales as $\mathcal{O}(p\cdot d^2)$. \emph{OperatorProbing} addresses this issue by introducing a lightweight pre-screening phase on a small subsample of the data, retaining only the most promising operators for the full pipeline.

Following the OpenFE-compatible cross-fitted candidate-selection protocol \cite{ref7}, we treat the original training and validation partitions as development data with distinct roles. We independently draw probe subsets at a ratio of $r_{\mathrm{probe}}$ from $D_{\mathrm{tr}}$ and $D_{\mathrm{sel}}$, where $D_{\mathrm{sel}}$ denotes the internal selection partition corresponding to the original OpenFE validation split. For classification tasks, stratified sampling is used when the class counts and requested subset size permit it; otherwise, uniform sampling without replacement is used. We compute three-fold OOF baseline predictions on the union $D_{\mathrm{probe}}^{\mathrm{tr}}\cup D_{\mathrm{probe}}^{\mathrm{sel}}$. Each observation is excluded from the fitting rows for the fold model that produces its OOF prediction; the held-out fold is used for early stopping and may contain observations from the internal selection partition. We therefore do not treat $D_{\mathrm{sel}}$ as a completely held-out model-evaluation set. The OOF predictions corresponding to $D_{\mathrm{probe}}^{\mathrm{sel}}$ are used to obtain the baseline loss $l_{\mathrm{init}}$. Candidate correction models are fitted on $D_{\mathrm{probe}}^{\mathrm{tr}}$ and evaluated on $D_{\mathrm{probe}}^{\mathrm{sel}}$, which is also used for early stopping. The test partition is never used for FeatureClustering, OperatorProbing, candidate scoring, feature attribution, or model selection.

For each operator $o\in O$, we randomly construct $n_{\mathrm{cand}}$ candidate features by applying $o$ to sampled features. The improvement of each candidate, $\Delta(\phi_j)$, is evaluated using the FeatureBoost mechanism defined in Eq.~\eqref{eq:featureboost}. By considering only the top-$k$ improvements, we prevent strong signals from being diluted by uninformative random pairs. The operator score is computed as follows:
\begin{equation}
\operatorname{Score}(o)=\frac{1}{k}\sum \Delta(\phi_j),
\qquad j\in\operatorname{Top}\text{-}k(o).
\label{eq:operator_score}
\end{equation}
The final selected operator set $O^*$ consists of the $N_{\mathrm{top}}$ highest-scoring operators according to $\operatorname{Score}(o)$. Algorithm~\ref{alg:operator_probing} describes the complete \emph{OperatorProbing} procedure.

\begin{algorithm}[t]
    \caption{Operator Probing}
    \label{alg:operator_probing}
    \begin{algorithmic}
        \REQUIRE $D_{\mathrm{tr}}$: training partition, $D_{\mathrm{sel}}$: internal candidate-selection partition (the original OpenFE validation split), $T=\{t_1,\ldots,t_d\}$: original features, $O=\{o_1,\ldots,o_p\}$: operator set \\
        \hspace*{1.5em} $r_{\mathrm{probe}}$: subsample ratio, $n_{\mathrm{cand}}$: candidates per operator, \\
        \hspace*{1.5em} $k$: top-$k$ scoring parameter, $N_{\mathrm{top}}$: number of operators to select
        \ENSURE $O^*\subseteq O$: selected operator subset

        \STATE $D_{\mathrm{probe}}^{\mathrm{tr}}\leftarrow\text{subsample}(D_{\mathrm{tr}},r_{\mathrm{probe}})$
        \STATE $D_{\mathrm{probe}}^{\mathrm{sel}}\leftarrow\text{subsample}(D_{\mathrm{sel}},r_{\mathrm{probe}})$
        \STATE $D_{\mathrm{probe}}\leftarrow D_{\mathrm{probe}}^{\mathrm{tr}}\cup D_{\mathrm{probe}}^{\mathrm{sel}}$
        \STATE $\hat{y}_{\mathrm{OOF}}\leftarrow\text{ThreeFoldCrossFit}(D_{\mathrm{probe}},T)$, excluding each observation from its fold's fitting rows and using each held-out fold for early stopping
        \STATE $l_{\mathrm{init}}\leftarrow\text{Loss}(y_{\mathrm{probe}}^{\mathrm{sel}},\hat{y}_{\mathrm{OOF}}^{\mathrm{sel}})$ on the internal selection observations
        \STATE \textbf{for each} $o\in O$:
        \STATE \hspace*{1.5em} $\Phi_o\leftarrow$ type-aware random sampling $(o,T,n_{\mathrm{cand}})$
        \STATE \hspace*{1.5em} \textbf{for each} $\phi_j\in\Phi_o$:
        \STATE \hspace*{3em} $\Delta(\phi_j)\leftarrow\mathrm{FeatureBoost}(D_{\mathrm{probe}}^{\mathrm{tr}},D_{\mathrm{probe}}^{\mathrm{sel}},\{\phi_j\},\hat{y}_{\mathrm{OOF}})$
        \STATE \hspace*{1.5em} $\mathrm{Score}(o)\leftarrow\mathrm{Mean}\bigl(\mathrm{Top}\text{-}k\{\Delta(\phi_j)\}\bigr)$
        \STATE $O^*\leftarrow$ top-$N_{\mathrm{top}}$ operators by Score
        \STATE \textbf{return} $O^*$
    \end{algorithmic}
\end{algorithm}

\subsection{Constrained Candidate Generation and Selection}
\label{subsec:constrained_generation}

Given the hard cluster assignment $C_{\mathrm{hard}}$ and selected operator set $O^*$, SCOPE-FE constructs the reduced pool $A=\mathcal{A}(T,O^*,C_{\mathrm{hard}})$. For every selected unary operator, candidates are generated from all type-compatible original features. For every selected two-input operator, candidates are generated only from type-compatible feature pairs satisfying $I_{\mathrm{hard}}(t_i,t_j)=1$. Candidate enumeration follows the original OpenFE implementation: numerical--numerical and categorical--categorical pairs are enumerated once under its operator-specific canonical ordering, while group-by operators use the fixed numerical-value/categorical-key argument order.

After candidate generation, SCOPE-FE does not introduce a new downstream selector. The standard OpenFE procedure computes baseline predictions, evaluates the reduced candidates using FeatureBoost-based successive halving, estimates feature attribution for the surviving candidates, and retains at most $k_{\mathrm{sel}}$ generated features. Consequently, SCOPE-FE changes where computation is allocated during expansion while preserving the downstream evaluation and selection protocol.

\subsection{Search-Space Effect}
\label{subsec:search_space_effect}

The exact reduction produced by FeatureClustering depends on the cluster-size distribution. Let $n_k=|C_k|$ and consider the underlying unordered feature-pair space. The number of intra-cluster pairs retained by the hard partition is
\begin{equation}
|\mathcal{P}_{C_{\mathrm{hard}}}|=\sum_{k=1}^{K}\binom{n_k}{2},
\label{eq:intra_pair_count}
\end{equation}
whereas the unconstrained pair space contains $\binom{d}{2}$ pairs. Dividing the count in Eq.~\eqref{eq:intra_pair_count} by the unconstrained count gives the pair-space retention ratio
\begin{equation}
\rho_{\mathrm{pair}}(C_{\mathrm{hard}})=\frac{\sum_{k=1}^{K}\binom{n_k}{2}}{\binom{d}{2}}.
\label{eq:pair_retention}
\end{equation}
This expression accounts for potentially imbalanced agglomerative clusters. When clusters are approximately balanced, $n_k\approx d/K$, yielding $\rho_{\mathrm{pair}}(C_{\mathrm{hard}})\approx 1/K\approx\tau/d$ and an intra-cluster pair count of approximately $\mathcal{O}(d\tau)$. This balanced-cluster expression is an approximation rather than a worst-case guarantee; the exact reduction is determined by Eq.~\eqref{eq:pair_retention}.

Let $p_2=|O_2|$ and $p_2^*=|O_2^*|$ denote the numbers of original and selected two-input operators. Ignoring differences in operator-specific type compatibility, the binary-candidate retention ratio can be approximated as
\begin{equation}
\frac{|A_2|}{|\mathcal{U}_2|}\approx\frac{p_2^*}{p_2}\,\rho_{\mathrm{pair}}(C_{\mathrm{hard}}).
\label{eq:combined_retention}
\end{equation}
Equation~\eqref{eq:combined_retention} shows how FeatureClustering and OperatorProbing act on complementary dimensions of the candidate space: FeatureClustering serves as a structural pair gate that reduces the number of admissible feature pairs, while OperatorProbing provides dataset-specific utility control over the operators applied to those pairs. Together they provide joint quality-aware candidate-space control. The empirical raw counts reported in the main text reflect type compatibility, validity, and argument-ordering rules but precede canonicalization and deduplication. Canonicalized unique counts used in the enrichment analysis are reported separately in Appendix~\ref{app:enrichment_details}.

\begin{table*}[t]
\caption{Overall predictive performance over 10 seeds (mean $\pm$ std). CA/MI/ME use RMSE, TE/BR/DI/NO/VE use ROC-AUC, and JA/CO use accuracy. Bold denotes the best value at the reported precision; ``--'' indicates unsupported tasks and ``$\times$'' a 24-hour timeout.}
\label{tab:main_perf}
\centering
\scriptsize
\resizebox{\textwidth}{!}{%
\begin{tabular}{lcccccccccc}
\toprule
Method & CA $\downarrow$ & MI $\downarrow$ & ME $\downarrow$ & TE $\uparrow$ & BR $\uparrow$ & DI $\uparrow$ & NO $\uparrow$ & VE $\uparrow$ & JA $\uparrow$ & CO $\uparrow$ \\
\midrule
Base
& 0.433$\pm$0.003 & 0.744$\pm$0.000 & 1128.8$\pm$1.639 & 0.669$\pm$0.001 & 0.747$\pm$0.005 & 0.731$\pm$0.001 & \textbf{0.996$\pm$0.000} & 0.925$\pm$0.000 & 0.721$\pm$0.002 & 0.968$\pm$0.001 \\
FCTree
& 0.443$\pm$0.002 & 0.744$\pm$0.000 & 1074.4$\pm$1.215 & 0.670$\pm$0.001 & 0.747$\pm$0.006 & 0.731$\pm$0.002 & 0.995$\pm$0.001 & 0.926$\pm$0.000 & 0.721$\pm$0.001 & 0.971$\pm$0.000 \\
SAFE
& -- & -- & -- & 0.674$\pm$0.001 & 0.749$\pm$0.004 & 0.730$\pm$0.002 & \textbf{0.996$\pm$0.000} & 0.925$\pm$0.000 & -- & -- \\
AutoCross
& -- & -- & -- & 0.646$\pm$0.001 & 0.764$\pm$0.001 & 0.728$\pm$0.001 & 0.994$\pm$0.000 & 0.921$\pm$0.000 & -- & -- \\
FETCH
& 0.437$\pm$0.002 & $\times$ & 1098.8$\pm$1.110 & 0.670$\pm$0.001 & $\times$ & 0.732$\pm$0.002 & 0.995$\pm$0.000 & 0.927$\pm$0.000 & 0.721$\pm$0.001 & $\times$ \\
OpenFE
& \textbf{0.421$\pm$0.002} & \textbf{0.738$\pm$0.000} & \textbf{981.3$\pm$1.745} & \textbf{0.681$\pm$0.002} & 0.785$\pm$0.002 & 0.888$\pm$0.002 & \textbf{0.996$\pm$0.000} & \textbf{0.928$\pm$0.000} & \textbf{0.729$\pm$0.001} & \textbf{0.974$\pm$0.000} \\
SCOPE-FE
& 0.429$\pm$0.001 & \textbf{0.738$\pm$0.000} & 985.3$\pm$2.1 & 0.671$\pm$0.002 & \textbf{0.787$\pm$0.002} & \textbf{0.890$\pm$0.001} & 0.995$\pm$0.000 & \textbf{0.928$\pm$0.000} & \textbf{0.729$\pm$0.001} & \textbf{0.974$\pm$0.000} \\
\bottomrule
\end{tabular}
}
\end{table*}

\begin{table*}[t]
\caption{Raw candidate counts and feature-engineering RUN. Counts precede canonicalization and deduplication; SCOPE-FE RUN is component-summed and includes FeatureClustering overhead. Reduction and speedup follow Eqs.~\eqref{eq:retention_reduction} and \eqref{eq:speedup_def}.}
\label{tab:candidate_runtime}
\centering
\scriptsize
\resizebox{\textwidth}{!}{%
\begin{tabular}{lrrrrrrrrrrl}
\toprule
Measure & CA & MI & ME & TE & BR & DI & NO & VE & JA & CO & Summary \\
\midrule
OpenFE candidates (raw)
& 275 & 77{,}643 & 346 & 13{,}882 & 21{,}047 & 7{,}772 & 81{,}044 & 30{,}500 & 9{,}018 & 27{,}333 & -- \\
SCOPE-FE candidates (raw)
& 132 & 12{,}541 & 78 & 2{,}892 & 2{,}516 & 918 & 9{,}163 & 5{,}068 & 1{,}590 & 8{,}571 & -- \\
Reduction
& 52.0\% & 83.8\% & 77.5\% & 79.2\% & 88.0\% & 88.2\% & 88.7\% & 83.4\% & 82.4\% & 68.6\% & Median 82.9\% \\
\midrule
OpenFE RUN (s)
& 34.90 & 4250.17 & 111.80 & 171.10 & 2143.40 & 218.68 & 306.50 & 215.72 & 326.98 & 1979.32 & -- \\
SCOPE-FE RUN (s)
& \textbf{18.82} & \textbf{978.94} & \textbf{56.05} & \textbf{70.36} & \textbf{852.50} & \textbf{83.06} & \textbf{55.91} & \textbf{91.99} & \textbf{106.48} & \textbf{1160.03} & -- \\
Speedup
& 1.85$\times$ & 4.34$\times$ & 1.99$\times$ & 2.43$\times$ & 2.51$\times$ & 2.63$\times$ & 5.48$\times$ & 2.35$\times$ & 3.07$\times$ & 1.71$\times$ & GM 2.66$\times$ \\
\bottomrule
\end{tabular}
}
\end{table*}

\section{Experiments}
\label{sec:experiments}

\subsection{Experimental Setup}
\label{subsec:experimental_setup}

\subsubsection{Datasets}
\label{subsubsec:datasets}
We evaluate SCOPE-FE on the same ten benchmark datasets used in the OpenFE evaluation suite \cite{ref7}: three regression tasks (California Housing, Microsoft, Medical), five binary-classification tasks (Telecom, Broken Machine, Diabetes, Nomao, VehicleNorm), and two multiclass-classification tasks (Jannis, 4 classes; Covtype, 7 classes). In tables and figures we abbreviate these as CA, MI, ME, TE, BR, DI, NO, VE, JA, and CO, respectively. We retain the original OpenFE train/validation/test split names for reproducibility. Operationally, the training partition $D_{\mathrm{tr}}$ and the original validation split, denoted $D_{\mathrm{sel}}$, form the development data used for candidate selection. FeatureClustering constructs its association matrix from $D_{\mathrm{tr}}$ only. OperatorProbing uses cross-fitted subsamples from both development partitions under the protocol described in Section~\ref{subsec:operator_space}, and the standard OpenFE successive-halving and feature-attribution stages also use both partitions. The held-out test partition remains untouched throughout feature-space construction and model selection and is used only for final predictive evaluation. Full dataset statistics and feature counts are given in Appendix~\ref{app:dataset_hparams}.

\subsubsection{Compared Methods}
\label{subsubsec:compared_methods}
We compare two groups of methods that serve different purposes and are therefore not mixed in the same table. \textit{General AutoFE methods:} the original feature set (Base), FCTree \cite{ref15}, SAFE \cite{ref16}, AutoCross \cite{ref17}, FETCH \cite{ref3}, and OpenFE \cite{ref7}, evaluated under the same fixed splits and metrics as SCOPE-FE. SAFE and AutoCross are applicable only to binary-classification tasks (marked ``--'' elsewhere), and FETCH runs exceeding the 24-hour limit are reported as timeouts. AutoFeat is mentioned in Section~\ref{sec:related_work} but omitted from Table~\ref{tab:main_perf}, as complete, same-protocol results across all ten datasets are not available for it.

\textit{SCOPE-FE diagnostic variants:} FC-only (hard FeatureClustering only, all operators retained), OP-only (OperatorProbing only, full pair space), SCOPE-FE (hard FeatureClustering + OperatorProbing, the full method), Random-Pair (selected operators fixed, pair selection randomized at matched binary-candidate budget), Random-Operator (Hard-Intra pairs fixed, operator assignment randomized at matched budget), Random-Joint (both pairs and operators randomized at matched budget), and the Hard-Inter diagnostic reported in Appendix~\ref{app:intra_inter}.

\subsubsection{Implementation and Repetition}
\label{subsubsec:implementation}
The downstream learner is LightGBM \cite{ke2017lightgbm}. Candidate scoring, successive halving, and feature attribution follow OpenFE's standard FeatureBoost-based pipeline unmodified; SCOPE-FE only changes the candidate pool $A$ that this pipeline operates on. For each algorithm and dataset, the downstream LightGBM evaluation is repeated over 10 random seeds, and we report the mean and standard deviation of the test-set metric across seeds. Candidate generation and scoring use $\tau=16$ and $N_{\mathrm{top}}=7$; the complete configuration, operator space, and hardware details are listed in Appendix~\ref{app:dataset_hparams}, Appendix~\ref{app:operator_space}, and Appendix~\ref{app:software_env}. Code and reproducibility materials will be publicly released upon acceptance.

\subsection{Evaluation Metrics}
\label{sec:candidate_quality_metrics}

\subsubsection{Predictive Performance}
Regression datasets are evaluated with RMSE (lower is better); binary-classification datasets with ROC-AUC (higher is better); multiclass-classification datasets with accuracy (higher is better). Because seed variance can be small enough that even a negligible mean difference tests as statistically significant, we use a metric-specific practical-equivalence margin as a descriptive criterion for comparing SCOPE-FE against OpenFE: an absolute margin of $0.005$ for ROC-AUC and accuracy, and a relative margin of $1\%$ for RMSE. The practical-equivalence margin is used as a post-hoc descriptive criterion rather than as a preregistered inferential threshold. It is intended to represent a small practical change relative to the reported scale of each metric, and we additionally provide the exact per-dataset differences so that the conclusions do not depend solely on the categorical win--tie--loss summary. A dataset is classified as a win, tie, or loss for SCOPE-FE according to whether the performance difference exceeds, falls within, or falls below the corresponding practical-equivalence margin.

\subsubsection{Candidate-Space Reduction}
Let $\mathcal{U}=\mathcal{A}(T,O)$ denote the full candidate universe (Eq.~\eqref{eq:full_candidate_universe}) and $A=\mathcal{A}(T,O^*,C_{\mathrm{hard}})$ the candidate pool actually generated by SCOPE-FE (Eq.~\eqref{eq:scope_candidate_space}). We report
\begin{equation}
\mathrm{Retention}=\frac{|A|}{|\mathcal{U}|}, \qquad
\mathrm{Reduction}=1-\frac{|A|}{|\mathcal{U}|}.
\label{eq:retention_reduction}
\end{equation}
Here $|\mathcal{U}|$ and $|A|$ (reported in Table~\ref{tab:candidate_runtime}) count the \emph{as-generated} candidates produced directly by the enumeration procedure, prior to the canonicalization and deduplication used for the enrichment analysis of Section~\ref{sec:candidate_quality_metrics}; the two counts differ by a small, dataset-dependent amount, detailed in Appendix~\ref{app:enrichment_details}.

\subsubsection{Runtime}
We use RUN as the primary efficiency metric. It covers feature engineering and excludes downstream LightGBM evaluation. OpenFE RUN is directly timed. SCOPE-FE RUN is component-summed. It adds FeatureClustering's separately measured same-machine cost to the timed stage spanning OperatorProbing through feature attribution. It is therefore reported explicitly as component-summed RUN rather than as a directly observed end-to-end wall-clock measurement. The two measured intervals are disjoint and, by construction, cover all feature-engineering stages exactly once; no stage is omitted or double-counted. The resulting decomposition is given in Appendix~\ref{app:runtime_breakdown}. Given $\mathrm{RUN}_{\text{OpenFE}}$ and $\mathrm{RUN}_{\text{SCOPE-FE}}$ for a dataset,
\begin{equation}
\mathrm{Speedup}=\frac{\mathrm{RUN}_{\text{OpenFE}}}{\mathrm{RUN}_{\text{SCOPE-FE}}}.
\label{eq:speedup_def}
\end{equation}
Per-seed evaluation time (Eval) and the resulting total time (RUN + Eval) are reported in Appendix~\ref{app:runtime_breakdown}; the abstract and main text are stated in terms of RUN throughout. Because RUN excludes downstream evaluation, all efficiency claims in the main text are restricted to the feature-engineering stage.

\subsubsection{Candidate-Pool Quality Metrics}
To test whether $A$ is not merely smaller but preferentially retains high-utility candidates, we rank the full candidate universe $\mathcal{U}$ by the exhaustively computed individual FeatureBoost gain $\Delta(\phi)$ (Eq.~\eqref{eq:featureboost}) and measure how much of the globally top-ranked slice SCOPE-FE's pool recovers. With $N=|\mathcal{U}|$, $M=|A|$, and $H=\bigl|A\cap\mathrm{Top}_M(\mathcal{U};\Delta)\bigr|$,
\begin{equation}
\begin{aligned}
\mathrm{Coverage}&=\frac{H}{M}, \qquad
\mathrm{ExpectedCoverage}_{\text{random}}=\frac{M}{N},\\
\mathrm{Enrichment}&=\frac{H/M}{M/N}.
\end{aligned}
\label{eq:enrichment_def}
\end{equation}
An enrichment greater than 1 indicates that $A$ contains a higher concentration of top-gain candidates than an equally sized subset drawn uniformly at random from $\mathcal{U}$ would be expected to. Significance is assessed with a one-sided hypergeometric test per dataset, with Benjamini--Hochberg false-discovery-rate correction \cite{benjamini1995controlling} across the ten datasets. In the main text we report enrichment as the primary candidate-quality metric; coverage, $H$, and raw/adjusted $p$-values are given in Appendix~\ref{app:enrichment_details}. Because the audit ranks candidates using the same FeatureBoost gain employed by the downstream OpenFE pipeline, enrichment should be interpreted as within-protocol candidate utility rather than model-agnostic interaction quality.

\begin{figure*}[t]
    \centering
    \begin{subfigure}[b]{0.47\linewidth}
        \centering
        \includegraphics[width=\linewidth]{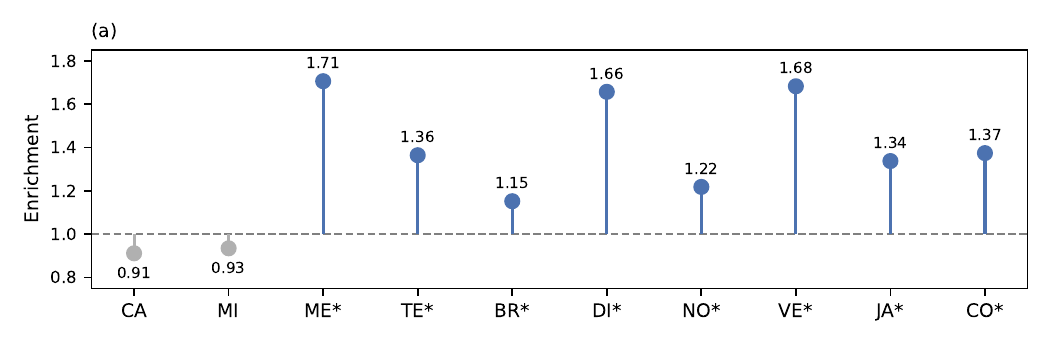}
        \caption{Exhaustive candidate-pool enrichment.}
        \label{fig:enrichment_quality_dotplot}
    \end{subfigure}
    \hfill
    \begin{subfigure}[b]{0.47\linewidth}
        \centering
        \includegraphics[width=\linewidth]{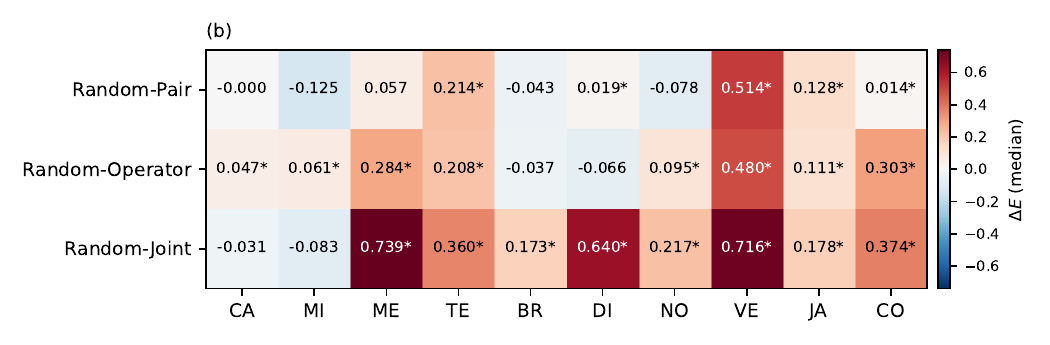}
        \caption{Budget-matched randomization controls.}
        \label{fig:random_control_heatmap}
    \end{subfigure}
    \caption{Candidate-pool quality. (a) Exhaustive candidate audit; the dashed line marks uniform-random expectation ($E=1$). (b) Median enrichment gap against budget-matched randomization controls over 1{,}000 seeds. Asterisks indicate BH-adjusted $q<0.05$.}
    \Description{
    Two-panel figure showing dataset-level candidate-pool
    enrichment and enrichment differences between SCOPE-FE and three budget-matched randomized controls.
    }
    \label{fig:candidate_pool_quality}
\end{figure*}

\subsection{Overall Predictive Performance and Efficiency}
\label{subsec:main_results}

We first summarize efficiency. SCOPE-FE decreases component-summed RUN on all ten datasets, achieving a geometric-mean speedup of $2.66\times$ and a maximum speedup of $5.48\times$ on Nomao (Table~\ref{tab:candidate_runtime}), alongside a median candidate-space reduction of $82.9\%$. These feature-engineering runtime figures add FeatureClustering's separately measured same-machine cost to the timed remaining stages, as detailed in Appendix~\ref{app:fc_overhead}.

Under the practical-equivalence margin defined in Section~\ref{sec:candidate_quality_metrics}, SCOPE-FE is $0$W~/~$8$T~/~$2$L against OpenFE (Table~\ref{tab:main_perf}): it remains practically comparable to OpenFE on 8 of 10 datasets. California Housing and Telecom exceed the practical-equivalence margin and are reported as losses rather than omitted: on California Housing, SCOPE-FE's RMSE is $2.0\%$ higher than OpenFE's (above the $1\%$ margin), and on Telecom, SCOPE-FE's ROC-AUC is $0.010$ lower than OpenFE's (above the $0.005$ margin). Appendix~\ref{app:margin_sensitivity} reports the margin-sensitivity analysis.

Taken together, these results indicate that SCOPE-FE primarily improves computational efficiency rather than uniformly increasing predictive performance; on California Housing and Telecom, where the candidate-space restriction appears to exclude candidates relevant to downstream performance, the resulting performance loss is visible rather than hidden by the aggregate statistics.

\subsection{Candidate-Pool Quality}
\label{subsec:candidate_pool_quality}
We next test whether the reduced pool preferentially retains high-utility candidates; full results are reported in Appendix~\ref{app:candidate_pool_quality}.

\subsubsection{Exhaustive Utility Enrichment}
\label{subsubsec:exhaustive_enrichment}
Using an exhaustive candidate audit of the full universe (per-candidate $\Delta(\phi)$ computed for every $\phi\in\mathcal{U}$, not only the retained $\phi\in A$), enrichment (Eq.~\eqref{eq:enrichment_def}) exceeds uniform-random expectation on 8 of 10 datasets, with a median enrichment of $1.35\times$ (mean $1.33\times$, range $0.91$--$1.71\times$), as shown in Figure~\ref{fig:candidate_pool_quality}\subref{fig:enrichment_quality_dotplot}. The enrichment remains significant after Benjamini--Hochberg correction on 8 of 10 datasets ($q<0.05$); the two exceptions, California Housing and Microsoft, are also the two datasets with enrichment below 1. Full per-dataset values ($N$, $M$, $H$, coverage, raw and adjusted $p$-values) are given in Appendix~\ref{app:enrichment_details}.

\subsubsection{Randomization Controls}
\label{subsubsec:random_controls}
Across 1{,}000 budget-matched randomizations, SCOPE-FE achieves higher median enrichment on 6 of 10 datasets against Random-Pair and 8 of 10 against both Random-Operator and Random-Joint, with BH-adjusted significance on 5, 8, and 8 datasets, respectively (Figure~\ref{fig:candidate_pool_quality}\subref{fig:random_control_heatmap}). The weaker Random-Pair result suggests that FeatureClustering functions more consistently as a structural pair-space reducer than as a universally effective utility filter. Full per-dataset results and testing details are reported in Appendix~\ref{app:random_controls}.

\subsection{Mechanism Analysis}
\label{subsec:mechanism_analysis}

\subsubsection{Component Ablation}
\label{subsubsec:component_ablation}
Table~\ref{tab:ablation_agg} decomposes SCOPE-FE into FC-only and OP-only under the same evaluation protocol. Both components individually reduce the candidate space on all ten datasets. FC-only reduces RUN on all ten datasets, whereas OP-only reduces RUN on eight and is slower than OpenFE on Medical and VehicleNorm. Their combination (SCOPE-FE) yields the largest aggregate candidate-space reduction and geometric-mean speedup of the three. FeatureClustering serves primarily as a deterministic structural pair gate, whereas OperatorProbing supplies dataset-specific utility control; SCOPE-FE combines these roles into joint quality-aware candidate-space control. Their predictive effects remain dataset-dependent: SCOPE-FE obtains slightly higher mean performance than OpenFE on Broken Machine and Diabetes, but both differences remain within the defined practical-equivalence margin, so we do not characterize either component as uniformly improving predictive performance or candidate quality. Their main benefit is the stronger aggregate candidate-space reduction and lower feature-engineering runtime achieved by jointly controlling the pair and operator spaces. Full per-dataset predictive values for all four legs are given in Appendix~\ref{app:full_ablation}. 

As an additional diagnostic, we compare intra- and inter-cluster pairing in Appendix~\ref{app:intra_inter}; the results favor Hard-Intra in aggregate but do not imply that useful interactions occur only within clusters.

\begin{table}[t]
\caption{Component ablation across ten datasets. Reduction and RUN speedup are relative to OpenFE; ``Comparable'' denotes practical equivalence.}
\label{tab:ablation_agg}
\centering
\small
\begin{tabular*}{\columnwidth}{@{\extracolsep{\fill}}lccccc@{}}
\toprule
Variant  & Pair & Op. & Med.\ red. & GM speedup & Comparable/10 \\
\midrule
OpenFE   & --   & --   & 0.0\%  & 1.00$\times$ & -- \\
FC-only  & Yes  & --   & 37.8\% & 1.90$\times$ & 8 \\
OP-only  & --   & Yes  & 66.5\% & 1.44$\times$ & 8 \\
SCOPE-FE & Yes  & Yes  & 82.9\% & 2.66$\times$ & 8 \\
\bottomrule
\end{tabular*}
\end{table}

\section{Conclusion}
\label{sec:conclusion}

We presented SCOPE-FE, a pre-generation candidate-space control framework that combines a structural pair gate with dataset-specific operator control while leaving the downstream OpenFE evaluation pipeline unchanged. Across ten OpenFE benchmarks, SCOPE-FE reduced the median candidate space by $82.9\%$, decreased component-summed RUN on every dataset, and achieved a geometric-mean speedup of $2.66\times$, while remaining within the stated practical-equivalence margins of OpenFE on 8 of 10 datasets. Exhaustive candidate auditing and budget-matched randomization controls further showed that the retained candidate pool is utility-enriched on most evaluated datasets and that this enrichment is not explained by candidate count alone. Because these experiments reproduce OpenFE's cross-fitted development-pool selection protocol, the results should not be interpreted as performance under a train-only probing regime with a completely held-out selection partition. Future work should examine stricter train-only or nested selection protocols and adaptive scope control across additional AutoFE pipelines, downstream learners, and benchmark settings.

\bibliographystyle{ACM-Reference-Format}
\bibliography{scope-fe_reference}

\appendix

\section{Notation}
\label{app:notation}

Table~\ref{tab:notation} summarizes the notation used in the formulation, algorithms, and candidate-pool analysis.

\begin{table}[H]
\caption{Notation used throughout the paper.}
\label{tab:notation}
\centering
\scriptsize
\setlength{\tabcolsep}{3pt}
\begin{tabular}{@{}>{\footnotesize}p{0.24\columnwidth}p{0.70\columnwidth}@{}}
\toprule
Symbol & Description \\
\midrule
$D$ & Supervised dataset $\{(\mathbf{x}_i,y_i)\}_{i=1}^{n}$ \\
$D_{\mathrm{tr}}$ & Training partition within the OpenFE-compatible development protocol \\
$D_{\mathrm{sel}}$ & Internal candidate-selection partition corresponding to the original OpenFE validation split \\
$n$ & Number of observations \\
$T=\{t_1,\ldots,t_d\}$ & Original feature set, with $d=|T|$ \\
$O=\{o_1,\ldots,o_p\}$ & Full operator set, with $p=|O|$ \\
$O_1,O_2$ & One-input and two-input operator subsets \\
$\mathcal{A}(T,O)$ & Valid candidates generated from features $T$ and operators $O$ \\
$\mathcal{U}$ & Full candidate universe $\mathcal{A}(T,O)$ \\
$\mathcal{B}$ & Arbitrary candidate pool, $\mathcal{B}\subseteq\mathcal{U}$ \\
$A$ & Reduced SCOPE-FE candidate pool \\
$A_{\mathrm{selected}},A_{\mathrm{final}}$ & Pools after successive halving and feature attribution \\
$O^*$ & Operator subset selected by OperatorProbing \\
$C_{\mathrm{hard}}$ & Hard cluster-assignment function $T\rightarrow\{1,\ldots,K\}$ \\
$C_k$ & Set of features assigned to cluster $k$ \\
$K$ & Number of clusters, $\max(2,\lceil d/\tau\rceil)$ \\
$\tau$ & Target cluster size \\
$n_k=|C_k|$ & Number of features in cluster $C_k$ \\
$I_{\mathrm{hard}}(t_i,t_j)$ & Indicator that $t_i$ and $t_j$ belong to the same cluster \\
$S,S_{ij}$ & Mixed-type feature-association matrix and its entries \\
$\widetilde{D}$ & Precomputed dissimilarity matrix, $\widetilde{D}_{ij}=1-S_{ij}$ \\
$\rho,V_{\mathrm{corr}},\eta^2$ & Pearson correlation, bias-corrected Cramér's V, and eta-squared \\
$L$ & Downstream learning algorithm \\
$E$ & Validation utility oriented so that larger values are better \\
$P(\mathcal{B})$ & Ideal achievable downstream performance from pool $\mathcal{B}$ \\
$k_{\mathrm{sel}}$ & Maximum number of generated features retained \\
$\hat{y}$ & Baseline out-of-fold predictions \\
$\phi,\phi_j$ & Individual generated candidate feature \\
$\Delta(\phi_j)$ & FeatureBoost gain of candidate $\phi_j$ \\
$G$ & Pre-generation scope-control function \\
$\mathrm{Cost}(A)$ & Cost of constructing, generating, and evaluating $A$ \\
$\epsilon$ & Allowed downstream-utility difference in the idealized objective \\
$l_{\mathrm{init}}$ & Baseline loss on the internal-selection probe observations \\
$r_{\mathrm{probe}}$ & OperatorProbing subsample ratio \\
$n_{\mathrm{cand}}$ & Maximum sampled candidates per operator \\
$\Phi_o$ & Sampled probe-candidate set for operator $o$ \\
$k$ & Number of top candidate gains used to score each operator \\
$N_{\mathrm{top}}$ & Number of operators retained by OperatorProbing \\
$\rho_{\mathrm{pair}}(C_{\mathrm{hard}})$ & Fraction of unordered feature pairs retained by $C_{\mathrm{hard}}$ \\
$p_2,p_2^*$ & Numbers of full and selected two-input operators \\
$N,M$ & Canonicalized unique counts $|\mathcal{U}|$ and $|A|$ in the audit \\
$H$ & Number of retained candidates in the global top-$M$ set \\
$\mathrm{Top}_M(\mathcal{U};\Delta)$ & Top $M$ candidates in $\mathcal{U}$ ranked by gain \\
Coverage, Enrichment & Candidate-pool quality metrics defined in Eq.~\eqref{eq:enrichment_def} \\
RUN & Feature-engineering time; component-summed for SCOPE-FE \\
Speedup & $\mathrm{RUN}_{\mathrm{OpenFE}}/\mathrm{RUN}_{\mathrm{SCOPE-FE}}$ \\
$\Delta E$ & SCOPE-FE enrichment minus a randomized control's median enrichment \\
$q$ & Benjamini--Hochberg-adjusted $p$-value \\
\bottomrule
\end{tabular}
\end{table}

\FloatBarrier
\section{Experimental Details}
\label{app:dataset_hparams}

\subsection{Datasets and Splits}

Table~\ref{tab:datasets} reports the dataset sizes, feature counts, and evaluation metrics under the fixed splits used throughout the experiments.

\begin{table}[t]
\caption{Dataset characteristics under the fixed OpenFE train/validation/test splits. Feature counts correspond to the inputs used by the feature-engineering pipeline after preprocessing.}
\label{tab:datasets}
\centering
\small
\begin{tabular*}{\columnwidth}{@{\extracolsep{\fill}}lrrrl@{}}
\toprule
Dataset & Train & Val.\ / Test & Feat.\ & Metric \\
\midrule
CA (Regr.)      & 13{,}209  & 3{,}303 / 4{,}128     & 8   & RMSE \\
MI (Regr.)      & 723{,}412 & 235{,}259 / 241{,}521 & 136 & RMSE \\
ME (Regr.)      & 104{,}361 & 26{,}091 / 32{,}613   & 11  & RMSE \\
TE (Bin.\ cls.) & 32{,}669  & 8{,}168 / 10{,}210    & 57  & ROC-AUC \\
BR (Bin.\ cls.) & 576{,}000 & 144{,}000 / 180{,}000 & 58  & ROC-AUC \\
DI (Bin.\ cls.) & 65{,}129  & 16{,}283 / 20{,}354   & 47  & ROC-AUC \\
NO (Bin.\ cls.) & 22{,}465  & 6{,}000 / 6{,}000     & 118 & ROC-AUC \\
VE (Bin.\ cls.) & 60{,}000  & 18{,}528 / 20{,}000   & 100 & ROC-AUC \\
JA (Multi., 4)  & 53{,}588  & 13{,}398 / 16{,}747   & 54  & Accuracy \\
CO (Multi., 7)  & 371{,}847 & 92{,}962 / 116{,}203  & 54  & Accuracy \\
\bottomrule
\end{tabular*}
\end{table}

Full names: CA = California Housing, MI = Microsoft, ME = Medical, TE = Telecom, BR = Broken Machine, DI = Diabetes, NO = Nomao, VE = VehicleNorm, JA = Jannis, CO = Covtype.

\subsection{Implementation and Hyperparameters}

Table~\ref{tab:hparams} lists the SCOPE-FE hyperparameters shared across the reported experiments, together with the stated dataset-specific exception for the final selected-feature count.

\begin{table}[t]
\caption{SCOPE-FE hyperparameters used to produce all reported results.}
\label{tab:hparams}
\centering
\small
\begin{tabular*}{\columnwidth}{@{\extracolsep{\fill}}ll@{}}
\toprule
Hyperparameter & Value \\
\midrule
Target cluster size $\tau$                  & 16 \\
Selected operator count $N_{\mathrm{top}}$  & 7 \\
Probing subsample ratio $r_{\mathrm{probe}}$ & 0.2 \\
Sampled candidates per operator $n_{\mathrm{cand}}$ & 20 \\
Top-$k$ for operator score & 5 \\
Final selected feature count $k_{\mathrm{sel}}$ & 10 (50 for JA, CO) \\
Agglomerative linkage & Average \\
Distance metric & Precomputed (Eq.~\eqref{eq:hard_distance}) \\
Parallel workers $n_{\mathrm{jobs}}$ & 8 \\
Downstream evaluation seeds & 10 \\
\bottomrule
\end{tabular*}
\end{table}

OperatorProbing's internal subsampling uses a fixed seed (seed $=1$) for every dataset, so the top-7 operator selection reported in Appendix~\ref{app:operator_space} is deterministic and reproducible given the same data.

\subsection{Operator Search Space}
\label{app:operator_space}
Table~\ref{tab:operator_space} lists the full set of candidate transformation operators available to OpenFE and SCOPE-FE, grouped by the feature-type signature they apply to. OperatorProbing (Section~\ref{subsec:operator_space}) selects $N_{\mathrm{top}}=7$ of these per dataset; the selected set is \emph{dataset-specific}, not fixed globally. Table~\ref{tab:selected_operators} reports the full selected operator set for all ten datasets.

\begin{table}[t]
\caption{Full candidate operator set, by arity and applicable feature-type signature.}
\label{tab:operator_space}
\centering
\small
\begin{tabular*}{\columnwidth}{@{\extracolsep{\fill}}ll@{}}
\toprule
Type & Operators \\
\midrule
Unary (any type) & \texttt{freq} \\
Unary (numeric)  & \texttt{abs}, \texttt{log}, \texttt{sqrt}, \texttt{square}, \\
                 & \texttt{sigmoid}, \texttt{round}, \texttt{residual} \\
Binary (num--num) & \texttt{min}, \texttt{max}, \texttt{+}, \texttt{-}, \texttt{*}, \texttt{/} \\
Binary (cat--num) & \texttt{GroupByThenMin}, \texttt{GroupByThenMax}, \\
                  & \texttt{GroupByThenMean}, \texttt{GroupByThenMedian}, \\
                  & \texttt{GroupByThenStd}, \texttt{GroupByThenRank} \\
Binary (cat--cat) & \texttt{Combine}, \texttt{CombineThenFreq}, \\
                  & \texttt{GroupByThenNUnique} \\
\bottomrule
\end{tabular*}
\end{table}

\begin{table}[t]
\caption{OperatorProbing's selected operator set ($N_{\mathrm{top}}=7$), per dataset, using the fixed probing seed described above.}
\label{tab:selected_operators}
\centering
\small
\begin{tabular}{@{}p{0.10\columnwidth}p{0.84\columnwidth}@{}}
\toprule
Dataset & Selected operators \\
\midrule
CA & \texttt{+}, \texttt{min}, \texttt{-}, \texttt{GroupByThenRank}, \texttt{/}, \texttt{*}, \texttt{max} \\
MI & \texttt{freq}, \texttt{*}, \texttt{GroupByThenMin}, \texttt{+}, \texttt{round}, \texttt{GroupByThenMedian}, \texttt{residual} \\
ME & \texttt{-}, \texttt{/}, \texttt{GroupByThenRank}, \texttt{+}, \texttt{min}, \texttt{GroupByThenMean}, \texttt{*} \\
TE & \texttt{*}, \texttt{CombineThenFreq}, \texttt{freq}, \texttt{GroupByThenStd}, \texttt{abs}, \texttt{GroupByThenMean}, \texttt{log} \\
BR & \texttt{abs}, \texttt{round}, \texttt{max}, \texttt{-}, \texttt{+}, \texttt{/}, \texttt{min} \\
DI & \texttt{freq}, \texttt{max}, \texttt{*}, \texttt{/}, \texttt{+}, \texttt{GroupByThenStd}, \texttt{sqrt} \\
NO & \texttt{+}, \texttt{residual}, \texttt{/}, \texttt{GroupByThenRank}, \texttt{*}, \texttt{-}, \texttt{sigmoid} \\
VE & \texttt{log}, \texttt{/}, \texttt{sigmoid}, \texttt{round}, \texttt{abs}, \texttt{residual}, \texttt{min} \\
JA & \texttt{freq}, \texttt{/}, \texttt{square}, \texttt{round}, \texttt{sigmoid}, \texttt{-}, \texttt{abs} \\
CO & \texttt{+}, \texttt{Combine}, \texttt{freq}, \texttt{CombineThenFreq}, \texttt{GroupByThenRank}, \texttt{-}, \texttt{GroupByThenStd} \\
\bottomrule
\end{tabular}
\end{table}

\subsection{Software Environment and Hardware}
\label{app:software_env}
All reported experiments, including the predictive-performance results (Table~\ref{tab:main_perf}), the OP + OpenFE stage and OpenFE RUN/Eval times (Table~\ref{tab:candidate_runtime}, Appendix~\ref{app:runtime_breakdown}), the candidate-pool audits (Appendix~\ref{app:candidate_pool_quality}), and FeatureClustering's overhead (Appendix~\ref{app:fc_overhead}), were executed using Python 3.11.7, LightGBM 4.6.0, and scikit-learn 1.8.0 on a MacBook Air equipped with an Apple M4 processor and 24\,GB of memory.

\section{Similarity Measures and Implementation Details}
\label{app:similarity_measures}
This appendix gives the formal definitions of the three association statistics used to construct the similarity matrix $S$ in Eq.~\eqref{eq:mixed_similarity}: Pearson correlation in Eq.~\eqref{eq:app_pearson}, bias-corrected Cram\'er's V in Eqs.~\eqref{eq:app_cramer} and \eqref{eq:app_cramer_v}, and eta-squared in Eq.~\eqref{eq:app_eta2}. It also specifies the degenerate-case handling used in the actual implementation.

\textit{Numeric--numeric.} For two numeric features $t_i,t_j$, we use the absolute value of Pearson's correlation coefficient,
\begin{equation}
S(t_i,t_j)=|\rho(t_i,t_j)|=\left|\frac{\mathrm{Cov}(t_i,t_j)}{\sigma_{t_i}\sigma_{t_j}}\right|.
\label{eq:app_pearson}
\end{equation}

\textit{Categorical--categorical.} We use the bias-corrected Cram\'er's V statistic, which extends the chi-square test of independence to a normalized $[0,1]$ measure of association. With $r,k$ the contingency-table dimensions and $n$ the sample size,
\begin{equation}
\varphi^2=\frac{\chi^2}{n}, \qquad
\varphi_{\mathrm{corr}}^2=\max\!\left(0,\ \varphi^2-\frac{(k-1)(r-1)}{n-1}\right),
\label{eq:app_cramer}
\end{equation}
\begin{equation}
V_{\mathrm{corr}}
=
\sqrt{
\frac{\varphi_{\mathrm{corr}}^2}
{\min(k_{\mathrm{corr}}-1,\ r_{\mathrm{corr}}-1)}
},
\label{eq:app_cramer_v}
\end{equation}
where $r_{\mathrm{corr}}=r-\frac{(r-1)^2}{n-1}$ and $k_{\mathrm{corr}}=k-\frac{(k-1)^2}{n-1}$ are the corresponding bias-corrected dimension terms.

\textit{Categorical--numeric.} We use eta-squared, which quantifies the proportion of variance in the numeric feature explained by group membership defined by the categorical feature,
\begin{equation}
\eta^2=\frac{SS_{\mathrm{between}}}{SS_{\mathrm{total}}}=\frac{\sum_g n_g(\bar{x}_g-\bar{x})^2}{\sum_i (x_i-\bar{x})^2}.
\label{eq:app_eta2}
\end{equation}

\textit{Degenerate cases.} Missing values are removed pairwise before computing each association, and infinite values in numerical arguments are first converted to missing values. For Pearson correlation and eta-squared, similarity is set to zero when fewer than three valid observations remain. Undefined statistics are also mapped to zero, including zero-variance numerical variables, degenerate contingency tables, and zero total numerical variance. The resulting similarities are clipped to $[0,1]$; the derived dissimilarity matrix is likewise clipped to $[0,1]$ and its diagonal is set to zero.

\section{Full Efficiency Results}
\label{app:runtime_breakdown}
\subsection{Component-Summed Runtime Breakdown}
All runtime components were measured on the same Apple M4 environment (Appendix~\ref{app:software_env}). For SCOPE-FE, the reported RUN is component-summed: it includes FeatureClustering, OperatorProbing, constrained candidate generation, successive halving, and feature attribution, but is not a directly observed end-to-end wall-clock measurement. FeatureClustering and the remaining stages were timed separately and summed per dataset; Table~\ref{tab:fc_overhead} (Appendix~\ref{app:fc_overhead}) reports the decomposition. Table~\ref{tab:runtime_breakdown} reports RUN, downstream evaluation runtime (Eval), and their sum (Total), for OpenFE and SCOPE-FE. Eval denotes the separate downstream LightGBM evaluation over 10 seeds and is not included in RUN.

\begin{table}[t]
\caption{Runtime breakdown (seconds). SCOPE-FE RUN and Total include FeatureClustering overhead.}
\label{tab:runtime_breakdown}
\centering
\small
\begin{tabular*}{\columnwidth}{@{\extracolsep{\fill}}lrrr|rrr@{}}
\toprule
 & \multicolumn{3}{c|}{OpenFE} & \multicolumn{3}{c}{SCOPE-FE} \\
Dataset & RUN & Eval & Total & RUN & Eval & Total \\
\midrule
CA & 34.90   & 47   & 81.90   & 18.82   & 17   & 35.82 \\
MI & 4250.17 & 2153 & 6403.17 & 978.94  & 953  & 1931.94 \\
ME & 111.80  & 918  & 1029.80 & 56.05   & 363  & 419.05 \\
TE & 171.10  & 47   & 218.10  & 70.36   & 27   & 97.36 \\
BR & 2143.40 & 202  & 2345.40 & 852.50  & 298  & 1150.50 \\
DI & 218.68  & 103  & 321.68  & 83.06   & 128  & 211.06 \\
NO & 306.50  & 58   & 364.50  & 55.91   & 54   & 109.91 \\
VE & 215.72  & 54   & 269.72  & 91.99   & 49   & 140.99 \\
JA & 326.98  & 396  & 722.98  & 106.48  & 383  & 489.48 \\
CO & 1979.32 & 4032 & 6011.32 & 1160.03 & 5320 & 6480.03 \\
\bottomrule
\end{tabular*}
\end{table}

Although feature-engineering RUN is lower on all ten datasets, Total is higher on Covtype (6480.03 s versus 6011.32 s) because the separate ten-seed downstream evaluation dominates the overall cost. We therefore do not claim a universal reduction in total end-to-end experimental time.

\subsection{FeatureClustering Overhead}
\label{app:fc_overhead}
FeatureClustering's cost consists of building the mixed-type similarity matrix $S$ (Eq.~\eqref{eq:mixed_similarity}, one association statistic per feature pair) and running agglomerative clustering on the resulting distance matrix (Eq.~\eqref{eq:hard_distance}). We measured this cost by re-executing the actual FeatureClustering implementation (\path{clustering/make_cluster_id.py}) on each dataset's real training data with the same configuration used to produce the paper's results ($\tau=16$, ordinal features treated as numeric, average-linkage agglomerative clustering on the precomputed distance matrix), on the same Apple M4 machine as every other RUN figure in this paper, and confirmed that the resulting cluster counts exactly match those used to produce the reported results. FeatureClustering overhead is dataset-dependent and accounts for up to $11.7\%$ of the reported SCOPE-FE RUN, with the largest relative contribution observed on Nomao.

Table~\ref{tab:fc_overhead} reports the measured FeatureClustering overhead, the remaining-stage RUN from OperatorProbing through feature attribution, their component-summed total, OpenFE RUN, and the resulting speedup. The dominant similarity-matrix construction cost is quadratic in the number of original features $d$, since it evaluates all $\binom{d}{2}$ feature pairs; the measured overhead is accordingly largest on Microsoft ($d=136$, $39.4$s) and smallest on California Housing ($d=8$, $0.04$s). This overhead changes the reported speedups but does not reverse SCOPE-FE's ranking against OpenFE on any dataset.

\begin{table}[t]
\caption{FeatureClustering overhead and component-summed RUN by dataset (seconds).}
\label{tab:fc_overhead}
\centering
\footnotesize
\begin{tabular*}{\columnwidth}{@{\extracolsep{\fill}}lrrrrr@{}}
\toprule
Dataset & \shortstack{FC\\overhead} & \shortstack{OP + OpenFE\\stage} &
\shortstack{Component-\\summed RUN} & \shortstack{OpenFE\\RUN} & Speedup \\
\midrule
CA & 0.04  & 18.78   & 18.82   & 34.90   & 1.85$\times$ \\
MI & 39.39 & 939.54  & 978.94  & 4250.17 & 4.34$\times$ \\
ME & 1.78  & 54.28   & 56.05   & 111.80  & 1.99$\times$ \\
TE & 2.92  & 67.44   & 70.36   & 171.10  & 2.43$\times$ \\
BR & 11.37 & 841.13  & 852.50  & 2143.40 & 2.51$\times$ \\
DI & 5.98  & 77.07   & 83.06   & 218.68  & 2.63$\times$ \\
NO & 6.52  & 49.39   & 55.91   & 306.50  & 5.48$\times$ \\
VE & 3.00  & 88.99   & 91.99   & 215.72  & 2.35$\times$ \\
JA & 0.84  & 105.64  & 106.48  & 326.98  & 3.07$\times$ \\
CO & 3.58  & 1156.46 & 1160.03 & 1979.32 & 1.71$\times$ \\
\midrule
GM / max & -- & -- & -- & -- & \shortstack{GM 2.66$\times$\\(max 5.48$\times$)} \\
\bottomrule
\end{tabular*}
\end{table}

\section{Candidate-Pool Quality}
\label{app:candidate_pool_quality}
\subsection{Exhaustive Enrichment Results}
\label{app:enrichment_details}
Table~\ref{tab:enrichment_details} reports the full exhaustive-audit statistics underlying Section~\ref{subsubsec:exhaustive_enrichment}: universe size $N$, retained pool size $M$, overlap with the global top-$M$ gain ranking $H$, coverage, expected random coverage, enrichment, and the raw and Benjamini--Hochberg-adjusted hypergeometric $p$-values.

Table~\ref{tab:candidate_runtime} (main text) and Table~\ref{tab:ablation_efficiency} (below) report raw candidate records passed to the evaluator, whereas $N$ and $M$ here use canonicalized unique candidate identities. Because ordinal features are compatible with both the numeric and categorical processing paths, raw enumeration can contain duplicate \texttt{freq} records for those features; canonicalization collapses these records, producing the small, dataset-dependent differences (0--53 candidates) between the reported raw counts and the audited values of $N$ and $M$. The same canonicalization rule is applied to both sides of every enrichment comparison in this paper, so this does not affect the validity of the enrichment statistics themselves.

\begin{table}[t]
\caption{Exhaustive enrichment by dataset. $N$ and $M$ are canonicalized unique candidate counts.}
\label{tab:enrichment_details}
\centering
\footnotesize
\resizebox{\columnwidth}{!}{%
\begin{tabular}{@{}lrrrccccc@{}}
\toprule
Dataset & \shortstack{$N$\\(unique)} & \shortstack{$M$\\(unique)} & $H$ &
\shortstack{Cov-\\erage} & \shortstack{Exp.\\cov.} &
\shortstack{En-\\rich.} & \shortstack{Raw\\$p$} & \shortstack{Adj.\\$q$} \\
\midrule
CA & 274     & 132    & 58   & 0.439 & 0.482 & 0.91 & 0.930 & 1.000 \\
MI & 77{,}618 & 12{,}516 & 1{,}887 & 0.151 & 0.161 & 0.93 & 1.000 & 1.000 \\
ME & 346     & 78     & 30   & 0.385 & 0.225 & 1.71 & $2.0\times10^{-4}$ & $2.9\times10^{-4}$ \\
TE & 13{,}868 & 2{,}878  & 815  & 0.283 & 0.208 & 1.36 & $5.1\times10^{-28}$ & $1.7\times10^{-27}$ \\
BR & 21{,}020 & 2{,}516  & 347  & 0.138 & 0.120 & 1.15 & $1.7\times10^{-3}$ & $2.2\times10^{-3}$ \\
DI & 7{,}762  & 908    & 176  & 0.194 & 0.117 & 1.66 & $4.8\times10^{-13}$ & $9.5\times10^{-13}$ \\
NO & 80{,}991 & 9{,}163  & 1{,}263 & 0.138 & 0.113 & 1.22 & $6.1\times10^{-15}$ & $1.5\times10^{-14}$ \\
VE & 30{,}500 & 5{,}068  & 1{,}417 & 0.280 & 0.166 & 1.68 & $7.3\times10^{-112}$ & $3.7\times10^{-111}$ \\
JA & 9{,}018  & 1{,}590  & 375  & 0.236 & 0.176 & 1.34 & $1.7\times10^{-11}$ & $2.8\times10^{-11}$ \\
CO & 27{,}288 & 8{,}526  & 3{,}661 & 0.429 & 0.312 & 1.37 & $1.9\times10^{-169}$ & $1.9\times10^{-168}$ \\
\bottomrule
\end{tabular}%
}
\end{table}

\subsection{Budget-Matched Randomization Controls}
\label{app:random_controls}
For every (dataset, control) pair underlying Section~\ref{subsubsec:random_controls} and Figure~\ref{fig:candidate_pool_quality}(b), we define the median enrichment gap as $\Delta E=E_{\text{SCOPE-FE}}-\mathrm{median}_{s=1}^{1000}E_{\text{control}}^{(s)}$. Table~\ref{tab:random_control_full} reports this gap over 1{,}000 random seeds, whether SCOPE-FE's enrichment exceeds the control's median, the raw empirical $p$-value, and the Benjamini--Hochberg-adjusted $q$-value. The BH adjustment is computed once, jointly, across all 30 rows of this table (not separately per control or per dataset), so the ``BH-adj.\ $q$'' column directly reflects the correction reported in the main text. For all three controls, unary candidates are held fixed to SCOPE-FE's own realized unary set, and only the binary-candidate allocation is randomized; candidate identities follow the same canonicalization rules as the exhaustive candidate audit in Appendix~\ref{app:enrichment_details}, including operator-specific argument ordering and deduplication. Each control is evaluated over 1{,}000 independent seeds (seeds $0$--$999$) per dataset.

\begin{table}[t]
\caption{Budget-matched randomization controls over 1{,}000 seeds per dataset. $p$-values are empirical and BH correction is joint across all 30 tests; $0.000$ denotes $|\Delta E|<5\times10^{-4}$.}
\label{tab:random_control_full}
\centering
\small
\resizebox{\columnwidth}{!}{%
\begin{tabular}{llcccc}
\toprule
Dataset & Control & Median $\Delta E$ & $\Delta E>0$ & Emp.\ $p$ & BH-adj.\ $q$ \\
\midrule
CA & Random-Pair     & $0.000$  & No  & 0.644 & 0.840 \\
CA & Random-Operator & $+0.047$ & Yes & 0.017 & 0.025 \\
CA & Random-Joint    & $-0.031$ & No  & 0.787 & 0.931 \\
MI & Random-Pair     & $-0.125$ & No  & 1.000 & 1.000 \\
MI & Random-Operator & $+0.061$ & Yes & 0.001 & 0.002 \\
MI & Random-Joint    & $-0.083$ & No  & 1.000 & 1.000 \\
ME & Random-Pair     & $+0.057$ & Yes & 0.382 & 0.520 \\
ME & Random-Operator & $+0.284$ & Yes & 0.030 & 0.043 \\
ME & Random-Joint    & $+0.739$ & Yes & 0.002 & 0.003 \\
TE & Random-Pair     & $+0.214$ & Yes & 0.001 & 0.002 \\
TE & Random-Operator & $+0.208$ & Yes & 0.001 & 0.002 \\
TE & Random-Joint    & $+0.360$ & Yes & 0.001 & 0.002 \\
BR & Random-Pair     & $-0.043$ & No  & 0.807 & 0.931 \\
BR & Random-Operator & $-0.037$ & No  & 0.939 & 1.000 \\
BR & Random-Joint    & $+0.173$ & Yes & 0.001 & 0.002 \\
DI & Random-Pair     & $+0.019$ & Yes & 0.001 & 0.002 \\
DI & Random-Operator & $-0.066$ & No  & 0.790 & 0.931 \\
DI & Random-Joint    & $+0.640$ & Yes & 0.001 & 0.002 \\
NO & Random-Pair     & $-0.078$ & No  & 1.000 & 1.000 \\
NO & Random-Operator & $+0.095$ & Yes & 0.001 & 0.002 \\
NO & Random-Joint    & $+0.217$ & Yes & 0.001 & 0.002 \\
VE & Random-Pair     & $+0.514$ & Yes & 0.001 & 0.002 \\
VE & Random-Operator & $+0.480$ & Yes & 0.001 & 0.002 \\
VE & Random-Joint    & $+0.716$ & Yes & 0.001 & 0.002 \\
JA & Random-Pair     & $+0.128$ & Yes & 0.002 & 0.003 \\
JA & Random-Operator & $+0.111$ & Yes & 0.002 & 0.003 \\
JA & Random-Joint    & $+0.178$ & Yes & 0.001 & 0.002 \\
CO & Random-Pair     & $+0.014$ & Yes & 0.011 & 0.017 \\
CO & Random-Operator & $+0.303$ & Yes & 0.001 & 0.002 \\
CO & Random-Joint    & $+0.374$ & Yes & 0.001 & 0.002 \\
\bottomrule
\end{tabular}
}
\end{table}

\subsection{OperatorProbing Probe-Seed Sensitivity}
\label{app:probe_seed_stability}
We assess sensitivity to OperatorProbing's subsampling seed on eight datasets (Medical, Telecom, Broken Machine, Diabetes, Nomao, VehicleNorm, Jannis, and Covtype) using probe seeds 1--10. Hard-Intra clusters, data splits, and all other hyperparameters are held fixed. For each seed, we rerun OperatorProbing and gated candidate generation, but not downstream feature selection or predictive evaluation, and recompute exhaustive candidate-pool enrichment using the same canonicalized universe and FeatureBoost ranking as in Appendix~\ref{app:enrichment_details}. Table~\ref{tab:probe_seed_stability} reports the mean pairwise Jaccard similarity over the 45 pairs of selected top-7 operator sets, the coefficient of variation (CV; sample standard deviation divided by the mean) of the canonicalized candidate count, and enrichment stability. California Housing and Microsoft are not included in this completed sensitivity subset, so no probe-seed stability claim is made for them.

\begin{table}[t]
\caption{OperatorProbing sensitivity over ten probe seeds on eight datasets. ``$E>1$'' is the number of seeds whose retained pool exceeds uniform-random enrichment.}
\label{tab:probe_seed_stability}
\centering
\small
\resizebox{\columnwidth}{!}{%
\begin{tabular}{lcccccc}
\toprule
Dataset & Jaccard & Count CV & Mean $E$ & $E$ CV & $E$ range & $E>1$ \\
\midrule
ME & 0.430 & 0.202 & 1.202 & 0.395 & 0.732--1.990 & 5/10 \\
TE & 0.236 & 0.318 & 1.421 & 0.081 & 1.267--1.586 & 10/10 \\
BR & 0.280 & 0.226 & 1.155 & 0.236 & 0.652--1.534 & 8/10 \\
DI & 0.389 & 0.441 & 2.715 & 0.348 & 1.372--3.576 & 10/10 \\
NO & 0.322 & 0.312 & 1.210 & 0.030 & 1.154--1.254 & 10/10 \\
VE & 0.384 & 0.288 & 1.366 & 0.160 & 1.116--1.683 & 10/10 \\
JA & 0.424 & 0.224 & 1.247 & 0.121 & 1.053--1.574 & 10/10 \\
CO & 0.451 & 0.132 & 1.320 & 0.066 & 1.229--1.520 & 10/10 \\
\midrule
Median & 0.387 & 0.257 & 1.283 & 0.141 & -- & -- \\
\bottomrule
\end{tabular}
}
\end{table}

The operator sets are not invariant to the probe seed: mean pairwise Jaccard similarity ranges from 0.236 to 0.451, and candidate-count CV ranges from 0.132 to 0.441. Nevertheless, enrichment exceeds 1 in 73 of the 80 runs and in every seed for Telecom, Diabetes, Nomao, VehicleNorm, Jannis, and Covtype. Medical and Broken Machine are less stable, with enrichment above 1 in 5/10 and 8/10 runs, respectively. These results support a limited within-protocol robustness claim for candidate-pool enrichment on most datasets in this subset, but not seed invariance of the selected operators or candidate budget.

\section{Full Component Ablation Results}
\label{app:full_ablation}
\subsection{Predictive Performance}
Table~\ref{tab:full_ablation} reports per-dataset predictive performance (mean $\pm$ std over 10 seeds) for OpenFE and the three SCOPE-FE ablation legs underlying Section~\ref{subsubsec:component_ablation}.

\begin{table}[t]
\caption{Full component-ablation predictive results, per dataset.}
\label{tab:full_ablation}
\centering
\small
\resizebox{\columnwidth}{!}{%
\begin{tabular}{llcccc}
\toprule
Dataset & Metric & OpenFE & FC-only & OP-only & SCOPE-FE \\
\midrule
CA & RMSE     & 0.4206$\pm$0.0016   & 0.4278$\pm$0.0026 & 0.4427$\pm$0.0025 & 0.4291$\pm$0.0014 \\
MI & RMSE     & 0.7382$\pm$0.0003   & 0.7378$\pm$0.0002 & 0.7379$\pm$0.0003 & 0.7381$\pm$0.0003 \\
ME & RMSE     & 981.3$\pm$1.745     & 981.73$\pm$1.61   & 982.75$\pm$1.47   & 985.28$\pm$2.09 \\
TE & ROC-AUC  & 0.6815$\pm$0.0020   & 0.6805$\pm$0.0018 & 0.6712$\pm$0.0018 & 0.6712$\pm$0.0018 \\
BR & ROC-AUC  & 0.7855$\pm$0.0019   & 0.7727$\pm$0.0021 & 0.7874$\pm$0.0029 & 0.7871$\pm$0.0025 \\
DI & ROC-AUC  & 0.8878$\pm$0.0025   & 0.8878$\pm$0.0025 & 0.8899$\pm$0.0010 & 0.8899$\pm$0.0010 \\
NO & ROC-AUC  & 0.9958$\pm$0.0001   & 0.9966$\pm$0.0002 & 0.9956$\pm$0.0002 & 0.9954$\pm$0.0004 \\
VE & ROC-AUC  & 0.9281$\pm$0.0004   & 0.9272$\pm$0.0003 & 0.9273$\pm$0.0003 & 0.9275$\pm$0.0002 \\
JA & Accuracy & 0.7286$\pm$0.0014   & 0.7331$\pm$0.0006 & 0.7319$\pm$0.0012 & 0.7293$\pm$0.0007 \\
CO & Accuracy & 0.9736$\pm$0.0005   & 0.9735$\pm$0.0007 & 0.9742$\pm$0.0003 & 0.9743$\pm$0.0003 \\
\bottomrule
\end{tabular}
}
\end{table}

\subsection{Candidate-Space Reduction and Feature-Engineering Runtime}
Table~\ref{tab:ablation_efficiency} reports, for the same four legs, the as-generated candidate-pool size and the feature-engineering RUN time (same definition as Appendix~\ref{app:runtime_breakdown}) underlying the aggregate reduction/speedup numbers in Table~\ref{tab:ablation_agg}. FC-only restricts pairing only (full operator set); OP-only restricts the operator set only (full pair space); SCOPE-FE restricts both.

\begin{table}[t]
\caption{Candidate counts and RUN time (seconds) for the component ablations. FC-only and SCOPE-FE RUN include FeatureClustering overhead.}
\label{tab:ablation_efficiency}
\centering
\small
\resizebox{\columnwidth}{!}{%
\begin{tabular}{lrrrr|rrrr}
\toprule
 & \multicolumn{4}{c|}{Candidates} & \multicolumn{4}{c}{RUN (s)} \\
Dataset & OpenFE & FC-only & OP-only & SCOPE-FE & OpenFE & FC-only & OP-only & SCOPE-FE \\
\midrule
CA & 275     & 227    & 175    & 132   & 34.90   & 14.87   & 16.98   & 18.82 \\
MI & 77{,}643 & 38{,}045 & 25{,}543 & 12{,}541 & 4250.17 & 1327.86 & 2656.30 & 978.94 \\
ME & 346     & 286    & 110    & 78    & 111.80  & 54.10   & 158.95  & 56.05 \\
TE & 13{,}882 & 10{,}472 & 3{,}858  & 2{,}892 & 171.10  & 88.91   & 157.24  & 70.36 \\
BR & 21{,}047 & 6{,}319  & 8{,}381  & 2{,}516 & 2143.40 & 1399.88 & 1881.39 & 852.50 \\
DI & 7{,}772  & 7{,}276  & 944    & 918   & 218.68  & 171.62  & 63.47   & 83.06 \\
NO & 81{,}044 & 30{,}866 & 23{,}087 & 9{,}163  & 306.50  & 124.30  & 204.85  & 55.91 \\
VE & 30{,}500 & 14{,}504 & 10{,}400 & 5{,}068  & 215.72  & 95.08   & 235.60  & 91.99 \\
JA & 9{,}018  & 4{,}392  & 3{,}132  & 1{,}590  & 326.98  & 185.90  & 159.79  & 106.48 \\
CO & 27{,}333 & 24{,}171 & 9{,}711  & 8{,}571  & 1979.32 & 1859.62 & 1411.21 & 1160.03 \\
\bottomrule
\end{tabular}
}
\end{table}

The FC-only and SCOPE-FE RUN values above sum all components, including FeatureClustering's similarity-matrix construction and clustering cost per dataset, timed separately on the same machine (Appendix~\ref{app:runtime_breakdown}); OP-only and OpenFE do not use FeatureClustering and are unaffected. Comparing the two components' per-dataset candidate-count reduction ratios, OperatorProbing (OP-only) produces the larger reduction on nine of the ten datasets, including datasets with many original features such as Microsoft ($67.1\%$ vs.\ FC-only's $51.0\%$) and Diabetes ($87.9\%$ vs.\ FC-only's $6.4\%$); FeatureClustering (FC-only) produces the larger reduction only on Broken Machine ($70.0\%$ vs.\ OP-only's $60.2\%$). RUN time does not track candidate count perfectly leg-by-leg (e.g.\ OP-only is slower than OpenFE on Medical despite fewer candidates), reflecting that successive-halving and attribution cost depends on more than pool size alone; SCOPE-FE, combining both restrictions, achieves the lowest component-summed RUN among all four legs on 7 of 10 datasets and remains faster than OpenFE on all ten.

\section{Hard-Intra versus Hard-Inter}
\label{app:intra_inter}
\subsection{Downstream Performance and Statistical Tests}
Table~\ref{tab:intra_inter_full} reports per-seed significance tests for Hard-Intra versus Hard-Inter on their naturally induced candidate pools under the evaluated clustering structure: paired $t$-tests and Wilcoxon signed-rank tests, both corrected across the ten datasets using Holm's procedure \cite{holm1979simple}. The verdict requires both the Holm-corrected paired $t$-test and the Holm-corrected Wilcoxon signed-rank test to indicate significance at $\alpha=0.05$ in the same direction; if either test is non-significant, the dataset is reported as no significant difference (n.s.). Hard-Intra is significantly better on 7 of 10 datasets, Hard-Inter is significantly better on 1 of 10 (Telecom), and the remaining 2 (Microsoft, Nomao) show no significant difference. Lower values are better for RMSE (CA, MI, ME); higher values are better for ROC-AUC (TE, BR, DI, NO, VE) and accuracy (JA, CO).

\begin{table}[t]
\caption{Hard-Intra versus Hard-Inter predictive performance and Holm-corrected tests over 10 seeds.}
\label{tab:intra_inter_full}
\centering
\small
\resizebox{\columnwidth}{!}{%
\begin{tabular}{lccccc}
\toprule
Dataset & Intra & Inter & $p_{t}$ (Holm) & $p_{W}$ (Holm) & Verdict \\
\midrule
CA & 0.4291$\pm$0.0014 & 0.4408$\pm$0.0015 & $1.2\times10^{-6}$ & 0.0195 & Intra \\
MI & 0.7381$\pm$0.0003 & 0.7382$\pm$0.0002 & 0.579 & 0.557 & n.s. \\
ME & 985.28$\pm$2.09    & 1084.25$\pm$1.15  & $9.8\times10^{-16}$ & 0.0195 & Intra \\
TE & 0.6712$\pm$0.0018 & 0.6725$\pm$0.0017 & 0.0130 & 0.0234 & Inter \\
BR & 0.7871$\pm$0.0025 & 0.7604$\pm$0.0033 & $5.7\times10^{-8}$ & 0.0195 & Intra \\
DI & 0.8899$\pm$0.0010 & 0.8879$\pm$0.0019 & 0.0414 & 0.0410 & Intra \\
NO & 0.9954$\pm$0.0004 & 0.9958$\pm$0.0005 & 0.0566 & 0.0977 & n.s. \\
VE & 0.9275$\pm$0.0002 & 0.9253$\pm$0.0003 & $5.7\times10^{-8}$ & 0.0195 & Intra \\
JA & 0.7293$\pm$0.0007 & 0.7208$\pm$0.0013 & $1.4\times10^{-7}$ & 0.0195 & Intra \\
CO & 0.9743$\pm$0.0003 & 0.9618$\pm$0.0007 & $1.1\times10^{-10}$ & 0.0195 & Intra \\
\bottomrule
\end{tabular}
}
\end{table}

\subsection{Budget-Matched Enrichment}
Table~\ref{tab:intra_inter_enrichment} reports the per-dataset enrichment results (median Intra $1.410\times$ versus median Inter $1.141\times$, Intra higher on 7 of 10 datasets). Downstream performance (Table~\ref{tab:intra_inter_full}) uses the candidate pools naturally induced by each pairing rule, whereas the enrichment comparison here matches the binary-candidate budget between Intra and Inter within each dataset before computing enrichment, so that the comparison isolates pair-selection quality from pool size; the two tables answer related but distinct questions and should not be conflated. These results compare complementary rules under the evaluated clustering structure and do not imply that intra-cluster pairs are universally more informative.

\begin{table}[t]
\caption{Budget-matched enrichment for Hard-Intra and Hard-Inter. ``Matched budget'' is the shared binary-candidate count.}
\label{tab:intra_inter_enrichment}
\centering
\small
\begin{tabular*}{\columnwidth}{@{\extracolsep{\fill}}lrccc@{}}
\toprule
Dataset & Matched budget & Intra & Inter & Difference \\
\midrule
CA & 43     & 1.482 & 1.186 & $+0.296$ \\
MI & 12{,}108 & 0.935 & 1.158 & $-0.223$ \\
ME & 32     & 3.717 & 1.014 & $+2.703$ \\
TE & 966    & 1.149 & 0.569 & $+0.580$ \\
BR & 2{,}400  & 1.152 & 1.298 & $-0.146$ \\
DI & 26     & 3.148 & 2.099 & $+1.049$ \\
NO & 8{,}985  & 1.218 & 1.301 & $-0.083$ \\
VE & 4{,}568  & 1.683 & 0.731 & $+0.951$ \\
JA & 1{,}320  & 1.338 & 1.124 & $+0.214$ \\
CO & 1{,}140  & 1.857 & 0.919 & $+0.938$ \\
\bottomrule
\end{tabular*}
\end{table}

\section{Practical-Equivalence Margin Sensitivity}
\label{app:margin_sensitivity}

Table~\ref{tab:margin_sensitivity} evaluates whether the descriptive predictive-performance summary depends on the selected practical-equivalence margins. Classification margins are absolute changes in ROC-AUC or accuracy, whereas regression margins are relative to the OpenFE RMSE. ``Comparable'' counts wins and ties, i.e., datasets on which SCOPE-FE is not practically worse than OpenFE under the corresponding margins.

\begin{table}[t]
\caption{Sensitivity of the descriptive practical-equivalence summary to narrower and wider margins.}
\label{tab:margin_sensitivity}
\centering
\small
\begin{tabular*}{\columnwidth}{@{\extracolsep{\fill}}cccc@{}}
\toprule
Classification & Regression & W/T/L & Comparable \\
\midrule
$0.0025$ & $0.5\%$ & 0/8/2 & 8/10 \\
$0.0050$ & $1.0\%$ & 0/8/2 & 8/10 \\
$0.0100$ & $2.0\%$ & 0/8/2 & 8/10 \\
\bottomrule
\end{tabular*}
\end{table}

The aggregate conclusion is unchanged when both metric-specific margins are halved or doubled. At the widest setting, California Housing remains just outside the $2\%$ RMSE margin (a $2.02\%$ increase), and Telecom remains just outside the $0.010$ classification margin (a $0.01027$ ROC-AUC decrease). Thus, the reported 8-of-10 result is not specific to the single default margin, although this analysis remains descriptive rather than an inferential equivalence test.

\clearpage
\end{document}